\documentclass[lettersize,journal]{IEEEtran}
\usepackage{amsmath,amsfonts}
\usepackage{algorithmic}
\usepackage{algorithm}
\usepackage{array}
\usepackage[caption=false,font=normalsize,labelfont=sf,textfont=sf]{subfig}
\usepackage{textcomp}
\usepackage{stfloats}
\usepackage{url}
\usepackage{color}
\usepackage{verbatim}
\usepackage{graphicx}
\usepackage{cite}
\usepackage[colorlinks,linkcolor=blue,anchorcolor=blue, citecolor=blue]{hyperref}
\usepackage{booktabs}
\graphicspath{{Figure/}}
\usepackage[american]{babel}
\usepackage{microtype}

\usepackage{mathrsfs, colortbl}
\usepackage{colortbl}
\definecolor{secondcolor}{RGB}{220,230,240}
\definecolor{firstcolor}{RGB}{241,220,219}

\usepackage {cleveref}
\begin{document}

\title{Deep Unfolding Multi-modal Image Fusion Network via Attribution Analysis}

\author{Haowen Bai, Zixiang Zhao,~\IEEEmembership{Member,~IEEE}, Jiangshe Zhang, Baisong Jiang,\\ Lilun Deng, Yukun Cui, Shuang Xu, Chunxia Zhang,~\IEEEmembership{Member,~IEEE}

\thanks{
This work has been supported by the National Natural Science Foundation of China under Grant 12201497 and 12371512, Shaanxi Fundamental Science Research Project for Mathematics and Physics under Grant 22JSQ033, Guangdong Basic and Applied Basic Research Foundation under Grant 2023A1515011358, 2024A1515011851 \textit{(Corresponding author: Jiangshe Zhang).}
}
\thanks{
Haowen Bai, Jiangshe
Zhang, Baisong Jiang, Lilun Deng, Yukun Cui, Chunxia Zhang are with the School of Mathematics and Statistics, Xi’an Jiaotong University, Xi’an, Shaanxi 710049, China (E-mail:
hwbaii@stu.xjtu.edu.cn, jszhang@mail.xjtu.edu.cn, bsjiangz@stu.xjtu.edu.cn, statdll@stu.xjtu.edu.cn, cuiyukun@stu.xjtu.edu.cn, cxzhang@mail.xjtu.edu.cn).
}
\thanks{
Zixiang Zhao is with the Photogrammetry and Remote Sensing, ETH Z\"urich,  8093 Z\"urich, Switzerland (E-mail: zixiang.zhao@hotmail.com)
}
\thanks{
Shuang Xu is with the School of Mathematics and Statistics, Northwestern
Polytechnical University, Xi’an, Shaanxi 710072, China (E-mail: xs@nwpu.edu.cn)}

}

\maketitle

\begin{abstract}
Multi-modal image fusion synthesizes information from multiple sources into a single image, facilitating downstream tasks such as semantic segmentation. 
Current approaches primarily focus on acquiring informative fusion images at the visual display stratum through intricate mappings. 
Although some approaches attempt to jointly optimize image fusion and downstream tasks, these efforts often lack direct guidance or interaction, serving only to assist with a predefined fusion loss.
To address this, we propose an ``Unfolding Attribution Analysis Fusion network'' (UAAFusion), using attribution analysis to tailor fused images more effectively for semantic segmentation, enhancing the interaction between the fusion and segmentation.
Specifically, we utilize attribution analysis techniques to explore the contributions of semantic regions in the source images to task discrimination. At the same time, our fusion algorithm incorporates more beneficial features from the source images, thereby allowing the segmentation to guide the fusion process.
Our method constructs a model-driven unfolding network that uses optimization objectives derived from attribution analysis, with an attribution fusion loss calculated from the current state of the segmentation network. We also develop a new pathway function for attribution analysis, specifically tailored to the fusion tasks in our unfolding network. An attribution attention mechanism is integrated at each network stage, allowing the fusion network to prioritize areas and pixels crucial for high-level recognition tasks.
Additionally, to mitigate the information loss in traditional unfolding networks, a memory augmentation module is incorporated into our network to improve the information flow across various network layers. Extensive experiments demonstrate our method's superiority in image fusion and applicability to semantic segmentation. The code is available at~\url{https://github.com/HaowenBai/UAAFusion}.
\end{abstract}


\section{Introduction}

{
\IEEEPARstart{i}{nfrared} 
and visible image fusion ~\cite{tang2023datfuse,sun2022drone,park2023cross,liu2021learning,zhao2023tufusion,li2023ccafusion, DBLP:journals/corr/abs-2211-14461, yang2021infrared,liu2024task,liu2023paif}, abbreviated as IVIF, integrates images of the same scene captured by different sensors to obtain a fused image that combines the advantages and information of both infrared and visible images.
Characterized by their distinct wavelengths, infrared and visible images capture diverse types of information.
Infrared images capture the thermal signatures of objects, a feature largely independent of ambient lighting.
This capability allows infrared images to penetrate smoke or clouds, thus highlighting objects within an environment. However, it struggles to capture the texture details of a scene and is susceptible to noise interference. Conversely, visible images depict the brightness, color, and texture of objects, providing a more intuitive and realistic representation. Nevertheless, they are sensitive to lighting conditions and obstructions, which can obscure the distinction between an object and its background.
IVIF extracts valid information from images of both modalities, enhancing the comprehensiveness and clarity of the image data. It eliminates the inherent imaging flaws of individual sensors to produce a single high-quality image and demonstrates robustness against disturbances~\cite{DBLP:journals/inffus/MaML19,DBLP:journals/tcsv/LiuCR20,DBLP:conf/aaai/JingLDWDSW20,DBLP:journals/tci/XuJWLSZZ20,DBLP:journals/tgrs/XuALZZL20}.
Therefore, the combination of information in infrared and visible images is widely applied in autonomous driving~\cite{DBLP:journals/inffus/LiZHWC20,DBLP:journals/tcsv/MaikCSP07}, face recognition~\cite{DBLP:journals/inffus/MaCLH16}, and security applications~\cite{DBLP:conf/icip/LahoudS18}, among others.
}

The fused image encapsulates rich information that single-modal images cannot fully capture. 
Thus, multi-modal image fusion is often combined with downstream tasks to achieve superior performance compared to single-modal inputs. Applications include semantic segmentation~\cite{DBLP:conf/iros/HaWKUH17,DBLP:journals/inffus/TangYM22,DBLP:journals/corr/abs-2308-02097,li2024object}, object detection~\cite{DBLP:journals/inffus/CaoGHYCQ19,DBLP:conf/cvpr/LiuFHWLZL22}, object tracking~\cite{DBLP:conf/eccv/LiZHTW18}, and pedestrian recognition~\cite{DBLP:conf/cvpr/LuWLZLCY20}.
{While existing deep learning-based approaches can successfully generate high-quality fusion images, most approaches only consider evaluation metrics or visual quality, which may not directly benefit high-level vision tasks~\cite{DBLP:conf/iconip/HarisSU21,DBLP:conf/eccv/PeiHZLW18,DBLP:conf/cvpr/LiARWTJCZGC19}.
Utilizing semantic richness~\cite{DBLP:journals/tmm/ZhouWZML23} or perceptual loss~\cite{DBLP:journals/inffus/LiWK21,DBLP:journals/tip/LiW19,DBLP:journals/inffus/ZhangLSYZZ20} to guide the fusion process improves the quality of the fused images. However, these methods may overlook their practical applications in downstream tasks.}
The extraction and integration of desired features by these methods are still not transparent, and their benefits for downstream tasks are unproven. 
Furthermore, using high-level task losses to guide the fusion process is challenging due to the varying levels and objectives of tasks~\cite{zhao2023metafusion}.
Therefore, there is a need for a fusion method that can preserve or even enhance the features or information required by downstream tasks. This would make the fused images more suitable for these applications.
{Such an algorithm needs to accomplish two goals: 1) extracting and preserving beneficial information from source images for downstream tasks, and 2) focusing on and carefully handling critical regions responding to the downstream tasks.}

Research indicates that tasks like semantic segmentation, object detection, and image classification rely not on all pixels equally, but specifically on areas such as edges, textures, or image highlights~\cite{DBLP:conf/aaai/VinogradovaDM20,DBLP:journals/corr/abs-2211-12108}. 
This implies that features from different regions contribute variably to the performance of downstream tasks. 
Thus, identifying and focusing on these regions is crucial for resolving downstream tasks.
Attribution analysis~\cite{DBLP:conf/cvpr/ShenGTZ20,DBLP:conf/iclr/BauZSZTFT19,DBLP:journals/corr/SimonyanVZ13,DBLP:journals/corr/SpringenbergDBR14,DBLP:conf/eccv/ZeilerF14} involves analyzing network outputs to measure and identify the contributions of different regions in the input image to the network's decisions.
For example, it identifies the pixels that prompt a trained network to classify an input image as a specific category. 
{Leveraging this advantage, we incorporate attribution analysis techniques into the fusion framework. Specifically, using semantic segmentation as an example, we explore how the fusion process and high-level vision tasks can mutually benefit each other.
Firstly, we utilize attribution analysis to dynamically measure the contribution of features from the source images to the semantic segmentation task during the training process. 
Based on these measurements, we establish attribution optimization objectives. We then construct the network architecture and define the attribution fusion loss.
This helps preserve beneficial features from the source images, thereby aiding semantic segmentation.}

{To further enhance the downstream adaptability of fused images, this paper directs the network's focus to pixels relevant to downstream tasks. We leverage attribution analysis to construct a task-driven attention mechanism. Traditional attribution analysis methods calculate the incremental contribution of features in the target image, typically using a reference image~\cite{DBLP:conf/icml/SundararajanTY17} and computing along a specific change path. However, in fusion tasks, the fused image is obtained by merging two source images. This means it has two reference images, making traditional path functions unsuitable for image fusion tasks. To model the transition from the two source images to the final fused result, we introduce algorithm unfolding techniques~\cite{DBLP:conf/icml/GregorL10,DBLP:journals/pami/YangSLX20,DBLP:journals/tci/LiTGME20,DBLP:journals/tgrs/XuALZZL20}. These techniques extend the attribution optimization objective into a deep unfolding network. By converting multiple iterations of the optimization objective into multiple layers of a neural network~\cite{DBLP:journals/spm/MongaLE21,wang2020model}, we treat the output of each layer as an intermediate state in the change path of the fused image. This constructs an attribution analysis path function suitable for image fusion. Algorithm unfolding also establishes a link between network architecture and optimization objectives, enhancing both performance and interpretability, and enabling a comprehensive understanding of its mechanisms. In each network layer, attribution analysis uses the output from both the current and previous stages to measure the fused image's contribution. It directs the network to attentively handle important pixel-level areas.}

In this work, we propose a novel fusion approach based on attribution analysis. 
This method preserves features and information from source images that benefit semantic segmentation, thereby facilitating the discrimination of fused images for segmentation, as illustrated in Fig.~\ref{fig:AUIF_net}.
We measure the contributions of different pixels in the source images to semantic segmentation and establish an optimization objective. 
This objective utilizes algorithmic unfolding techniques to further structure the network, transforming multiple iterations into a multi-level neural network stack.
The optimization objectives are also used to constitute the attribution fusion loss. 
Additionally, based on attribution analysis, we establish a novel path function suitable for the unfolding network. We also propose a new attribution attention mechanism, which enhances the focus of the fusion network on pixels most beneficial to semantic segmentation. 
Furthermore, to tackle the information compression issue in unfolding networks, we introduce a \textit{Memory Augmentation Module}. Operating independently of the information extraction stream, this module enhances interactions between neighboring and non-neighboring stages~\cite{DBLP:journals/ijcv/ZhouYPRXC23,DBLP:conf/mm/SongCZ21}.
Our main contributions are briefly summarized as follows:

{(1) We develop a new fusion objective based on attribution analysis, aimed at preserving beneficial information from source images for semantic segmentation. This objective not only guides the construction of fusion loss but also directs the training of the fusion network. It effectively allows segmentation to steer the fusion process, thereby enhancing the task adaptability of the fused images.}

{(2) Based on the attribution optimization objective, we propose a novel image fusion network derived through algorithmic unfolding. This network integrates interpretability with performance, employing memory augmentation techniques to enhance information flow.}

{(3) We introduce a new attention mechanism, ``Attribution Attention,'' utilizing attribution analysis to pinpoint pixels and regions crucial for segmentation. Embedding this module into the proximal operators of the deep unfolding network enhances focus on key feature components.}

{(4) We conducted exhaustive comparative experiments on fusion and segmentation, achieving significant results. These outcomes validate the rationality and effectiveness of our proposed method.}

The structure of this paper is as follows: In Section~\ref{sec:2}, we briefly describe related work pertinent to our method. In Section~\ref{sec:3}, our proposed methodology is presented in detail. In Section~\ref{sec:4}, we demonstrate comparative experiments on fusion and segmentation. Conclusions are given in Section~\ref{sec:5}.

\section{RELATED WORK}\label{sec:2}
In this part, we review various image fusion algorithms for infrared and visible images, and briefly introduce the methods and applications of algorithm unfolding and attribution analysis.
\subsection{Image Fusion}
Recent image fusion techniques can be categorized into traditional methods and deep learning-based methods.
Traditional methods include multiscale transform methods~\cite{DBLP:journals/sigpro/LiuJWSD14,DBLP:journals/ijon/LiuMD17,DBLP:journals/isci/0019LLM020}, sparse representation methods~\cite{DBLP:journals/tip/LiWK20,DBLP:journals/spl/LiuCWW16}, and subspace methods~\cite{DBLP:conf/icip/CvejicLBC06}.
Multiscale transformation methods leverage the decomposition features of the source images across varying scales to facilitate fusion. 
These methods aptly preserve the details of images while maintaining global consistency.
The commonly used methods encompass pyramid transform~\cite{BULANON200912}, discrete cosine transform~\cite{JIN20181, DBLP:journals/mta/EKR19}, discrete wavelet~\cite{DBLP:journals/sigpro/LiuJWSD14}, shearlet~\cite{ 
DBLP:journals/ijon/LiuMD17}, nonsubsampled contourlet transform~\cite{XIANG201553} and bilateral filter~\cite{DBLP:journals/inffus/HuL12}.
Sparse representation is also widely used for feature extraction~\cite{DBLP:journals/tip/LiWK20,DBLP:journals/spl/LiuCWW16}, using the sparse basis in an overcomplete dictionary to represent the source image. 
Methods based on sparse representation effectively retain intricate and global structural information of images, exhibiting robustness against noise and distortion.
In addition, subspace-based methods such as independent component analysis~\cite{DBLP:conf/icip/CvejicLBC06}, principal component analysis and non-negative matrix factorization~\cite{DBLP:conf/icip/ZhangWMW04} are used for image fusion, yielding high-fidelity fused images.
However, manually designed feature extraction and fusion methods, such as maximum, mean, and $\ell_1$-norm, are increasingly complex and often fail to meet the evolving demands for speed and effectiveness in fusion.

Traditional methods typically necessitate the manual design of the fusion criterion, thereby imposing limitations on performance potential. 
In contrast, neural networks, renowned for their adeptness in feature extraction and learning, have yielded a plethora of high-performance techniques based on deep learning paradigms~\cite{DBLP:conf/aaai/Xu0LJG20,DBLP:journals/inffus/ZhangLSYZZ20}.
Specifically, autoencoder-based methods train autoencoders for feature extraction and reconstruction, employing either a manually designed fusion strategy or network to merge encoder-extracted features and then reconstruct them with the decoder~\cite{DBLP:journals/tip/LiW19,DBLP:journals/inffus/LiWK21,DBLP:journals/tim/LiWD20,DBLP:conf/ijcai/ZhaoXZLZL20,DBLP:journals/corr/abs-2211-14461}.
GAN-based fusion methods innovatively use adversarial training, with the discriminator ensuring the generator produces high-quality fusion images~\cite{DBLP:journals/inffus/MaYLLJ19,DBLP:journals/inffus/MaLYCGWJ20}. Subsequently, a series of GAN-based networks have been proposed, including dual-discriminator model~\cite{DBLP:journals/tip/MaXJMZ20}, semantic-supervised model~\cite{DBLP:journals/tmm/ZhouWZML23}, and more.
Similarly, algorithm unfolding-based fusion models are a crucial branch of deep fusion networks, enhancing optimization models via iterative algorithms to develop network structures~\cite{DBLP:journals/tcsv/ZhaoXZLZL22,DBLP:journals/pami/0002D21,DBLP:journals/corr/abs-2005-08448}, which expand optimization models through iterative optimization algorithms to obtain the network structures.
Researchers have also explored unified multi-modal image fusion frameworks. These include continuous learning from multiple tasks~\cite{DBLP:journals/pami/XuMJGL22}, learning from decomposition and compression~\cite{DBLP:journals/ijcv/ZhangM21}, and techniques for learning fusion from single images by masking, without needing paired images for training~\cite{DBLP:conf/eccv/LiangJLM22}.
Due to imaging or shooting limitations, fused source image pairs often misalign. Adding a pre-alignment module to the network can prevent such issues~\cite{DBLP:conf/eccv/HuangLFLZL22,DBLP:conf/ijcai/WangLFL22,DBLP:conf/cvpr/Xu0YLL22}.

Moreover, several studies have attempted to combine the fusion task with the downstream tasks~\cite{DBLP:conf/cvpr/LiuFHWLZL22}, both at the image level~\cite{DBLP:journals/inffus/TangYM22,DBLP:conf/cvpr/LiuFHWLZL22,DBLP:journals/tmm/ZhouWZML23} and at the feature level~\cite{zhao2023metafusion}. 
Specifically, using the loss function for the downstream task to guide the parameters of the fusion network~\cite{DBLP:journals/inffus/TangYM22}, as well as applying semantic information metrics~\cite{DBLP:journals/tmm/ZhouWZML23} and perceptual loss~\cite{DBLP:journals/inffus/LiWK21,DBLP:journals/tip/LiW19,DBLP:journals/inffus/ZhangLSYZZ20}, have proven to be effective.
Additionally, meta-learning has been employed to utilize object detection features to guide the fusion at the feature level ~\cite{zhao2023metafusion}.

\subsection{Algorithm Unfolding}

Algorithm unfolding is first applied to fast sparse coding techniques~\cite{DBLP:conf/icml/GregorL10}. This approach extends traditional iterative algorithms to deep neural networks, incorporating predefined hyperparameters in the end-to-end training process.
In this context, sparse coding inference acts as the neural network's feed-forward process, with each iterative step functioning as a network layer~\cite{DBLP:conf/cvpr/ZhaoZXLP22}.
This technique has also been widely used in low-level vision. Examples include unfolding for the gradient total variation regularization model in image deblurring~\cite{DBLP:journals/tci/LiTGME20}, various sparse coding models~\cite{deng2019deep,DBLP:journals/tip/MarivaniTCD20}, and employing the maximum a posteriori probability model in image super-resolution~\cite{yang2022memory,DBLP:conf/cvpr/ZhangGT20}. It is also applied in general image restoration tasks~\cite{DBLP:conf/cvpr/ZhangZGZ17,DBLP:journals/pami/DongWYSWL19,DBLP:conf/iccv/LiuPRS19,DBLP:journals/ijcv/ZhouYPRXC23}.

In image fusion, methods similar to the convolutional sparse coding approach employ deep unfolding to achieve fusion~\cite{DBLP:journals/pami/0002D21,DBLP:journals/corr/abs-2005-08448}. Xu et al.~\cite{DBLP:conf/cvpr/Xu0ZSL021} first apply the algorithmic unfolding technique to pan-sharpening by modeling and unfolding the optimization-based pan-sharpening problem. By unfolding two optimization models, Zhao et al.~\cite{DBLP:journals/tcsv/ZhaoXZLZL22} successfully decompose the low-frequency and high-frequency parts of the source images. These parts are then fused and the final fusion image is reconstructed. Li et al.~\cite{li2023lrrnet} decompose the source image into low-rank and salient parts, and fuse them via a low-rank representation-guided learnable model.

\subsection{Attribution Analysis}
Attribution analysis aims to identify meaningful image structures or features, facilitating model interpretation~\cite{DBLP:conf/cvpr/ShenGTZ20}, model understanding~\cite{DBLP:conf/iclr/BauZSZTFT19} and model visualization~\cite{DBLP:journals/corr/SimonyanVZ13,DBLP:journals/corr/SpringenbergDBR14,DBLP:conf/eccv/ZeilerF14}. 
Since attribution analysis identifies regions crucial for classification decisions, it also supports weakly supervised object localization~\cite{DBLP:conf/cvpr/ZhouKLOT16,DBLP:conf/iccv/SelvarajuCDVPB17,DBLP:journals/corr/abs-1805-11393}, which involves locating objects in images using only classification masks.

Various attribution methods have emerged in recent years. 
For an input image $I$ and a classification model $S$, the gradient of the input image $\operatorname{Grad}S(I)=\frac{\partial y_{c}}{\partial I}$ can be visualized to find pixels that significantly influence the network output~\cite{DBLP:journals/jmlr/BaehrensSHKHM10,DBLP:journals/corr/SimonyanVZ13}. Here, $y_c$ represents the classification model's output for class $c$, typically the class with the highest score, which is selected as the discriminative result by the network.
The element-wise product of the input and its gradient $I \odot \frac{\partial S(I)}{\partial I}$, addresses the gradient saturation issue~\cite{DBLP:conf/icml/SundararajanTY17}. 
Then, CAM~\cite{DBLP:conf/cvpr/ZhouKLOT16} and Grad-CAM~\cite{DBLP:conf/iccv/SelvarajuCDVPB17} use the gradient of the convolutional features in the last layer of the model to locate the regions and pixels that are most relevant to the output. 
Integrated Gradients (IG)~\cite{DBLP:conf/icml/SundararajanTY17} introduces a baseline image $I^{\prime}$ and integrates the gradient along a path that gradually transitions from the baseline image to the target image $\left(I-I^{\prime}\right) \cdot \int_0^1 \frac{\partial S\left(I^{\prime}+\alpha\left(I-I^{\prime}\right)\right)}{\partial I} d \alpha$ to obtain the attribution map. 
The gradient is the network output relative to the intermediate image of the path. Attribution analysis methods such as Grad-CAM~\cite{DBLP:conf/iccv/SelvarajuCDVPB17} used for classification can also be extended to pattern recognition tasks. 
The attribution map for a semantic segmentation network~\cite{DBLP:conf/aaai/VinogradovaDM20} is computed by replacing the class output $y_c$ in the classification model with $\sum_{(i, j) \in \mathcal{M}} y_{ij}^c$, analyzing pixels $(i,j)$ within set $\mathcal{M}$, where $\mathcal{M}$ is the set of pixels to analyze and $y_{i j}^c$ is the $c$-th class score on pixel $(i,j)$. 
Similarly, the objectivity and classification scores in the output of the object detection network are subjected to the attribution operator, and the detection network can be likewise analyzed~\cite{DBLP:journals/corr/abs-2211-12108}. 
For tasks like super-resolution, which do not involve classification, attribution analysis is performed by analyzing local gradients $D_{x y}(I)=\sum_{i \in[x, x+l], j \in[y, y+l]} \nabla_{i j} I$ in the generated images~\cite{DBLP:conf/cvpr/GuD21}, where $i,j$ are the coordinates and $l$ is the region size.

\begin{figure*}[!]
	\centering
	\includegraphics[width=\linewidth]{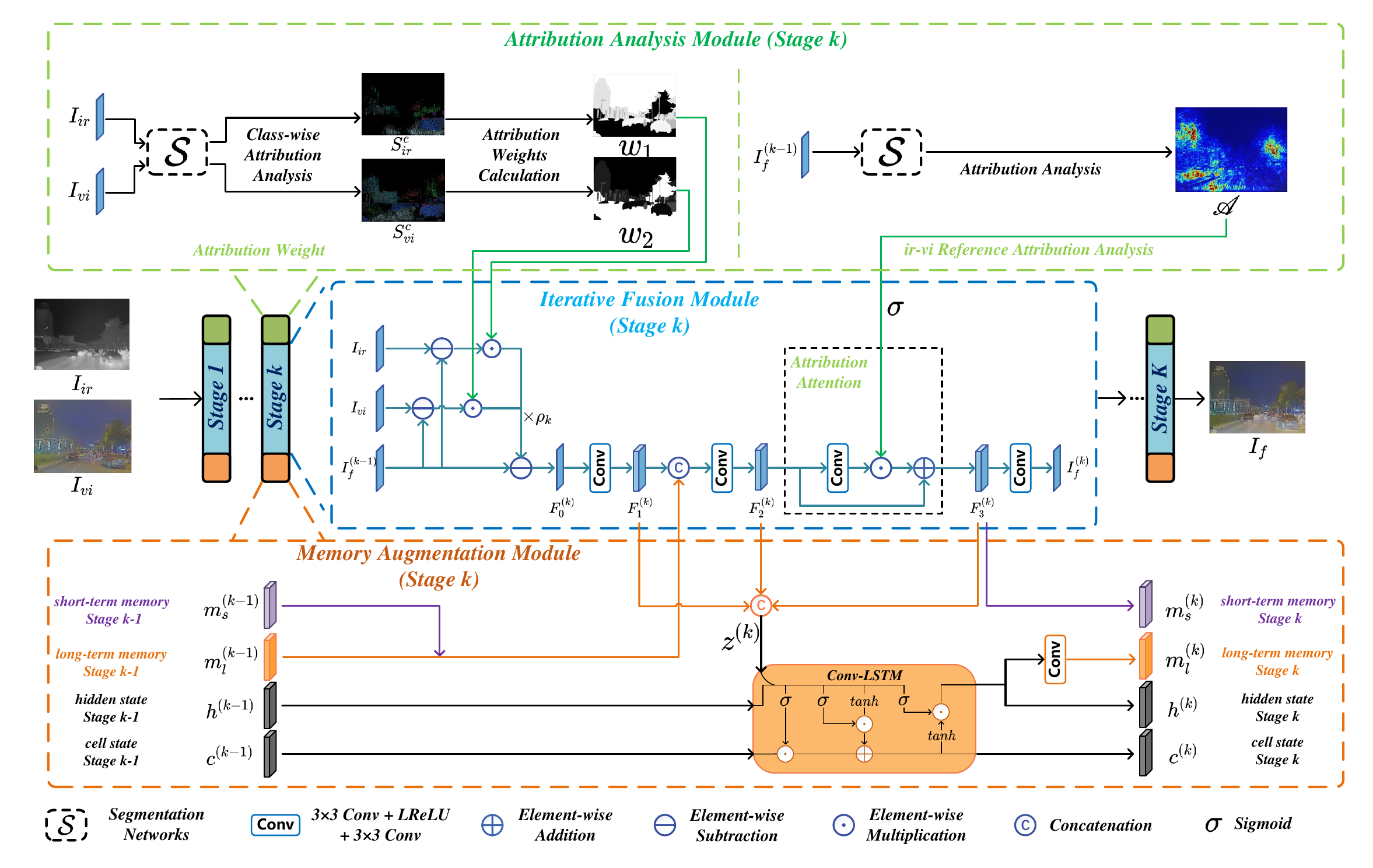}
	\caption{{Illustration of UAAFusion.} UAAFusion consists of multiple stages, each stage contains three parts, \textit{Iterative Fusion Module} (blue) for feature extraction as well as processing, the \textit{Attribution Analysis Module} (green) for calculating the attribution weight and attribution map, and the \textit{Memory Augmentation Module} (orange) for enhancing the flow of information between stages.}
	\label{fig:AUIF_net}
\end{figure*}

\section{Method}\label{sec:3}
In this section, we present our proposed methodology.
Prior to presenting our proposed network, we define the key notations used in this paper.
$I_{ir}$ and $ I_{vi}$ represent the infrared image and visible image, respectively. $I_{f}$ denotes the fused image, which retains the resolution of $I_{ir}$ and $I_{vi}$.

We design a downstream task-aware network using the algorithm unfolding technique.
Each stage of UAAFusion contains three important components, as shown in Fig.~\ref{fig:AUIF_net}. 
\begin{itemize}	
	\item The \textit{Iterative Fusion Module}, highlighted in blue, orchestrates the iterative fusion process, progressively refining the fusion outcomes. Each iteration of this module includes trainable convolutional layers for various functions such as channel expansion, prior memory integration, attribution attention focusing, and channel compression. The black dashed box labeled \textit{Attention Component} dynamically incorporates downstream task-specific attribution maps from the \textit{Attribution Analysis Module}, enabling targeted enhancement of critical features for task performance.
	
	\item The \textit{Attribution Analysis Module}, depicted in green, performs a detailed attribution analysis to assign weights to each semantic region of the source images. These weights not only guide the initial stages of image fusion in the \textit{Iterative Fusion Module} but also contribute to the computation of the fusion loss. It also generates stage-specific attribution maps for the fused image $I_f^{(k-1)}$, which are subsequently utilized by the \textit{Attention Component} to refine the focus on key image areas relevant to the segmentation.

	\item The \textit{Memory Augmentation Module}, represented in orange, manages both short-term and long-term memory across stages. It consolidates this memory data to prevent loss of information and ensure continuity between stages, thereby facilitating more effective interaction and integration of information throughout the network.
\end{itemize}

{Contrary to current image fusion methods that primarily focus on downstream tasks, we propose a new optimization objective. This objective is specifically designed to select information from source images that enhances downstream task discrimination, thereby producing fusion images with improved task adaptability. This approach allows the segmentation process to directly influence the fusion objectives and process, moving beyond merely contributing to a predetermined fusion loss. Our network architecture is crafted around this optimization objective, utilizing the algorithmic unfolding technique. Within this framework, traditional optimization steps are replaced by neural network layers, enhancing the model's efficiency and scalability. This substitution not only enhances transparency and interpretability but also ensures consistency between the network structure and the optimization objectives. Furthermore, to boost the adaptability of the fusion images, we integrate an attribution attention mechanism within the network, tailored to the fusion task. This mechanism focuses processing on areas crucial for semantic segmentation. To mitigate information loss in deep unfolding networks, we employ a memory augmentation component that bolsters the information flow between layers. In the following subsections, we will delve into detailed explanations of the mechanisms and operational principles of each module.}

\subsection{Iterative Fusion Module}\label{IFM}
To ensure that the fused image $I_{f}$ preserves the essential information from the source images while integrating features that support the segmentation, we first formulate the following optimization problem:
\begin{equation}
	\label{equ1}
	\min _{I_{f}} \frac{w_1}{2}\odot (I_{f}-I_{ir})^2+ \frac{w_2}{2}\odot (I_{f}-I_{vi})^2+h(I_{f}),
\end{equation}
where $\odot$ denotes element-wise multiplication for each term.
The first two terms in Eq.~(\ref{equ1}) serve as data fidelity terms, ensuring that the fused image $I_{f}$ remains consistent with each source image.
The weights $w_1$ and $w_2$ denote the contribution of each source image to the downstream task, with their calculation detailed in the subsequent section.
The term $h(\cdot)$ represents the priority function for the downstream task, emphasizing essential features in $I_{f}$ that enhance its relevance for subsequent processing.
This optimization model ensures that the fused image achieves pixel-level similarity with the source images, thus enhancing its effectiveness for the downstream task by incorporating relevant features.

To iteratively solve the optimization model and monitor changes in the fused image with respect to $I_{ir}$ and $I_{vi}$,  we employ the gradient projection method to solve Eq.~(\ref{equ1}):
\begin{equation}
	\label{equ2}
	I_{f}^{(k)}=\operatorname{prox}_h\left(I_{f}^{(k-1)}-\rho_k \nabla f\left(I_{f}^{(k-1)}\right)\right),
\end{equation}
where $k \in 1,2,3, \cdots K $, $\rho_k$ is the learnable step size, and $f(\cdot)$ is the objective function defined in Eq. (\ref{equ1}).
Here, $\operatorname{prox}_h(\cdot)$ denotes the proximal operator associated with the priority term $h(\cdot)$, and
\begin{equation}
    \label{equ3}
    \nabla f\left(I_{f}^{(k-1)}\right)=w_1\odot(I_{f}-I_{ir})+w_2\odot(I_{f}-I_{vi})
\end{equation}
denotes the gradient of $f(\cdot)$ with respect to $I_{f}^{(k-1)}$.
Leveraging the principle of algorithm unfolding, we can efficiently learn $\operatorname{prox}_h(\cdot)$ in Eq.~(\ref{equ2}) through a data-driven approach that incorporates convolutional neural networks (CNN).
This approach is termed the \textit{Unfolding Attribution Analysis Fusion Network} (UAAFusion).
Each stage of UAAFusion employs a CNN-based architecture to enable feature extraction, memory augmentation, and attention injection.
In the $k$-th stage of the network, detailed in the blue area of Fig.~\ref{fig:AUIF_net}, the implementation is divided into the following steps:

\begin{equation}\label{eq4}
	\begin{split}
		\begin{aligned}
			F_0^{(k)}&=\left(I_{f}^{(k-1)}-\rho_k \nabla f\left(I_{f}^{(k-1)}\right)\right),	\\
			F_1^{(k)}&=\operatorname{Conv}\left(F_0^{(k)}\right),	\\
			F_2^{(k)}&=\operatorname{Conv}\left(\operatorname{Cat}\left(F_1^{(k)}, m_s^{(k-1)}, m_l^{(k-1)}\right)\right),\\
			F_3^{(k)}&=\sigma (\mathscr{A}) * \operatorname{Conv}(F_2^{(k)})+F_2^{(k)},\\
			I_{f}^{(k)}&=\operatorname{Conv}\left(F_3^{(k)}\right),
		\end{aligned}
	\end{split}
\end{equation}
where $F_{i}^{(k)},i=0,1,2,3$ represents the intermediate feature maps. $m_s^{(k-1)}$ and $m_l^{(k-1)}$ represent short-term and long-term memory features passed from the previous stage, respectively, while $\mathscr{A}$ denotes the attribution map obtained through attribution analysis.
Here, $\sigma$ represents the $Sigmoid(\cdot)$ function, which normalizes the values in $\mathscr{A}$ to a $[0, 1]$ range.
The specific calculations for $m_s^{(k-1)}$, $m_l^{(k-1)}$, and $\mathscr{A}$ are provided in the subsequent subsections.
The calculations in Eq.~(\ref{eq4}) correspond to the blue section in Fig.~\ref{fig:AUIF_net}.

\subsection{Attribution Analysis Module} 
To integrate the fusion task with semantic segmentation for mutual enhancement, we use attribution analysis to build the fusion network, guide its learning, and utilize attention mechanisms to improve performance.

In our framework, attribution analysis serves a dual purpose. 
On the one hand, attribution analysis calculates the attribution weights of infrared and visible images, as discussed in the sections on network construction and loss function design.
This weighting enables the network to effectively learn features beneficial to segmentation. 
On the other hand, attribution analysis generates attention maps by applying it to the fused images at each stage of the network.
These attention maps help highlight the specific pixels and regions within the image that facilitate the execution of the downstream task.

\subsubsection{The attribution weights $w_1$, $w_2$} 
Unlike conventional fusion methods that assign fixed weights to all source images, our approach dynamically evaluates the contribution of source image features to segmentation, enabling the fused image to better represent the most beneficial source features.
{Inspired by~\cite{DBLP:journals/tmm/ZhouWZML23}, we attempt to assign weights to each semantic class of the source image pairs and perform fusion accordingly.
Specifically, we measure the attribution values for each semantic category in the original image and convert these into attribution weights between the two source images. 
In image classification, attribution analysis typically involves examining category scores $y^c$ to locate regions relevant to the $c$-th category within the images.
In our framework, we focus on the class scores for each semantic category $c$, following~\cite{DBLP:conf/aaai/VinogradovaDM20}, as output by the segmentation network at corresponding locations as follows:}
\begin{equation}
{Score}^c(I)=\frac{1}{N^c} \sum_{(i, j)\in c}\left(y_{i j}^c\right),
\end{equation}
where $y_{i j}^c$ is the score for the $c$-th class at position $(i, j)$ in the segmentation results for image $I$.
$(i, j)\in c$ means that the pixel at position $(i, j)$ belongs to the $c$-th semantic class.
$N^c$ represents the total number of pixels that belong to the $c$-th semantic class.
We use the integrated gradients method~\cite{DBLP:conf/icml/SundararajanTY17} to derive the class-wise attribution scores, calculated as follows:

\begin{equation}
S_{ir/vi}^c=\int_0^1 \frac{\partial {Score}^c(\gamma(\alpha))}{\partial \gamma(\alpha)} \times \frac{\partial \gamma(\alpha)}{\partial \alpha} d \alpha,
\end{equation}
where $\gamma(\alpha):[0,1] \mapsto \mathbb{R}^{H \times W}$ is the path function that transforms gradually from the base image $I^{\prime}$ to the target image $I$, where $\gamma(0)=I^{\prime}$ and $\gamma(1)=I$. In calculating the attribution weights, a zero image serves as the reference, so $\gamma(\alpha)=\alpha I_{ir/vi}$.
{The above formula represents the cumulative change in the score for the $c$-th semantic class as the input image transitions from a zero image to $I_{ir/vi}$. 
This cumulative change reflects how the emergence of specific features in the image contributes to the segmentation network’s discrimination.
Pixels with higher scores indicate a greater impact on the segmentation network’s discrimination as they appear, highlighting their importance for semantic segmentation.}
By discretizing the path into $M$ steps, we can simplify the calculation of this score as follows:
\begin{equation}
	S_{ir/vi}^c:=\sum_{j=1}^M \frac{\partial {Score}^c (\frac{j}{M}\cdot I_{ir/vi})}{\partial\left(\frac{j}{M}\cdot I_{ir/vi}\right)} \cdot \frac{I_{ir/vi}}{M}.
\end{equation}
Normalizing the attribution scores for the source images $I_{ir}$ and $I_{vi}$ allows us to compute their respective attribution weights.
For example, the attribution weight $w_1$ for $I_{ir}$ is calculated as follows:
\begin{equation}
	\small
w_1((i, j) \in c)=\frac{\operatorname{ReLU}\left(\sum_{(i,j) \in c} S_{i r}^c\right)}{\operatorname{ReLU}\left(\sum_{(i,j) \in c} S_{i r}^c\right)\!+\!\operatorname{ReLU}\left(\sum_{(i, j)\in c} S_{vi}^c\right)}.
\end{equation}
Accordingly, $w_2$ is derived from $w_2=1-w_1$.
In experiments, small values $\epsilon$ and $2\epsilon$ are added to the numerator and denominator, respectively, to maintain numerical stability.
The rationale for using ReLU (Rectified Linear Unit) is that if one source image contributes non-positive features to the downstream task, the fused image should rely entirely on the other source image.
This approach ensures that only positive contributions from source images are considered in the fusion, effectively ignoring irrelevant or non-beneficial features from the less relevant source.
The calculated attribution weights ${w_1,w_2}$ guide both the network structure design, as detailed in Sec.~\ref{IFM}, and the fusion loss function. This strategy enables the network to prioritize features from the most informative source images, optimizing the fusion process for semantic segmentation.

\subsubsection{The Attribution Attention $\mathscr{A}$} \label{sec:AA}
{Unlike the previous section, where attribution scores are computed for each semantic region by category, here attribution analysis examines the fusion results for the entire image. This approach helps evaluate pixel significance and construct spatial attention for the fused image.} Thus, we directly analyze the scores of all pixels within the fused image, as follows:
\begin{equation}
	{Score}(I)=\frac{1}{N} \sum_{c} \sum_{(i, j)\in c}\left(y_{i j}^c\right) .
\end{equation}
where $N$ represents the total number of pixels in image $I$.
Similar to the previous section, we use the integrated gradients method to compute the fused image’s attribution map.
\begin{equation}\label{equ6}
	\mathscr{A}(\gamma)=\int_0^1 \frac{\partial {Score}(\gamma(\alpha))}{\partial \gamma(\alpha)} \times \frac{\partial \gamma(\alpha)}{\partial \alpha} d \alpha,
\end{equation}
where $\mathscr{A}(\gamma)$ is the attribution map of $I_f$.
However, as the fused image combines infrared and visible images, identifying an appropriate reference image and path function $\gamma(\alpha)$ poses challenges.
Fortunately, our network is a deep unfolding architecture aligned with optimization objectives, allowing us to construct suitable paths for each stage, especially for the $k$-th stage, using intermediate network outputs.
Guided by attribution analysis and the deep unfolding architecture, we define a novel path function as follows:
\begin{equation}\label{equ7}
	\gamma(\alpha)\!=\!I_{f}^{(l)}+k\left(\alpha\!-\!\frac{l}{k}\right)\left(I_{f}^{(l+1)}\!-\!I_{f}^{(l)}\right),\alpha \in\left[\frac{l}{k}, \frac{l+1}{k}\right],
\end{equation}
where $l \in\{0,1, \ldots, k\}$. 
From Eq.~(\ref{equ7}), we derive that $\gamma(0)=I_{f}^{(0)}$,$\gamma(\frac{l}{k})=I_{f}^{(l)}$ and $\gamma(1)=I_{f}^{(k)}$. That is, $\gamma(\alpha)$ transitions from $I_{f}^{(0)}$ to $I_{f}^{(k)}$ along ${I_{f}^{(1)},I_{f}^{(2)}, \ldots,I_{f}^{(k-1)}}$.
In practice, we sample the gradient at the output of each stage ${I_{f}^{(0)}, I_{f}^{(1)}, \ldots, I_{f}^{(k)}}$ to estimate the gradient, approximating Eq.~(\ref{equ6}) as follows:
\begin{equation}\label{equ8}
	\mathscr{A}(\gamma)_{i}^{\text {approx }}\!:=\!\sum_{j=1}^k \frac{\partial D S\left(I_{f}^{(j)}\right)}{\partial (I_{f}^{(j)})_{i}} \cdot\left(I_{f}^{(j)}-I_{f}^{(j-1)}\right)_{i}.
\end{equation}
This construction satisfies the condition $\sum_i \mathscr{A}(\gamma)_i = D S\left(I_f^{(k)}\right) - D S\left(I_f^{(0)}\right)$. This restriction aligns with the guidelines for attribution analysis methods proposed by~\cite{DBLP:journals/ijgt/Friedman04, DBLP:conf/icml/SundararajanTY17}.

The computed attribution map $\mathscr{A}$ is processed through a $Sigmoid(\cdot)$ function and incorporated into each stage's iterative fusion module, as depicted in Fig.~\ref{fig:AUIF_net} and Eq.~(\ref{eq4}). This attention mechanism evaluates the importance of different areas within the fusion image for downstream task discrimination, enhancing the algorithm’s focus on these regions.

\subsection{Memory Augmentation Module}
Since the output of each stage, $I_{f}^{(k)}$, is a 1-channel feature representation that often results in information loss, we employ a memory augmentation approach to enhance information transmission.
The specific mechanism of memory augmentation is detailed in the \textit{Memory Augmentation Module} depicted in Fig.~\ref{fig:AUIF_net}. 
This module includes both short-term memory between adjacent stages and long-term memory spanning all stages. 
The features just before the final convolution layer in the proximal operator of the current stage serve as the short-term memory. 
This short-term memory is used in the next stage, as expressed by: 
\begin{equation}\label{equ9}  
	m_s^{(k)}=F_3^{(k)}.
\end{equation}
In each stage, all multi-channel features $\mathbf{z}^{(k)}=\operatorname{Cat}(F_1^{(k)},F_2^{(k)},F_3^{(k)}),$ are fed into a Convolutional LSTM ($\operatorname{ConvLSTM}$) for cross-stage long-term memory, balancing past and current states~\cite{DBLP:journals/ijcv/ZhouYPRXC23,DBLP:conf/mm/SongCZ21}:
\begin{equation}
	\begin{split}
		\begin{aligned}
			\left[\mathbf{h}^{(k)}, \mathbf{c}^{(k)}\right]&=\operatorname{ConvLSTM}\left(\mathbf{z}^{(k)},\left[\mathbf{h}^{(k-1)}, \mathbf{c}^{(k-1)}\right]\right),\\
			m_l^{(k)}&=\operatorname{Conv}(\mathbf{h}^{(k)}),
		\end{aligned}
	\end{split}
\end{equation}
where $F_1^{(k)},F_2^{(k)}$ and $F_3^{(k)}$ are the intermediate features of the $k$-th stage proximal operator. $\mathbf{h}^{(k)}$ and $\mathbf{c}^{(k)}$ are the hidden state and cell state of $\operatorname{ConvLSTM}$ in $k$-th stage, respectively. 

The architecture of $\operatorname{ConvLSTM}$ in $k$-th stage is:
\begin{equation}
\begin{aligned}
	  \mathbf{i}^{(k)}&=\sigma\left(\mathcal{W}_{z i} \odot  \mathbf{z}^{(k)}+\mathcal{W}_{h i} \odot  \mathbf{h}^{(k-1)}+b_i\right), \\
	  \mathbf{f}^{(k)}&=\sigma\left(\mathcal{W}_{z f} \odot  \mathbf{z}^{(k)}+\mathcal{W}_{h f} \odot  \mathbf{h}^{(k-1)}+b_f\right), \\
	  \mathbf{c}^{(k)}&=\mathbf{f}^{(k)} \odot  \mathbf{c}^{(k-1)}\\ 
                        &+\mathbf{i}^{(k)} \odot  \tanh \left(\mathcal{W}_{z c} \odot  \mathbf{z}^{(k)}\!+\!\mathcal{W}_{h c} \odot  \mathbf{h}^{(k-1)}\!+\!b_c\right), \\
        \mathbf{o}^{(k)}&=\sigma\left(\mathcal{W}_{z o} \odot  \mathbf{z}^{(k)}+\mathcal{W}_{h o} \odot  \mathbf{h}^{(k-1)}+b_o\right), \\
	\mathbf{h}^{(k)}&=\mathbf{o}^{(k)} \odot  \tanh \left(\mathbf{c}^{(k)}\right).
\end{aligned}
\end{equation}
The $\operatorname{ConvLSTM}$ structure selectively retains or discards information and states from previous stages. 
Its capability facilitates the efficient preservation of long-range stage correlations.

\subsection{Loss Function}
Our fusion framework consists of both a fusion network and a segmentation network, trained simultaneously using a synchronous training strategy.
To ensure that the fused image retains essential information from the source images, the training process uses the following loss function:  
\begin{equation}\label{equ10}  
	\mathcal{L}_{total}=\mathcal{L}_{int}+\lambda \mathcal{L}_{grad}+\mu \mathcal{L}_{seg}.  
\end{equation}  
The intensity loss enhances segmentation-relevant information from the source images, while the gradient loss maximizes texture and detail preservation from the originals.
The segmentation loss simultaneously trains the segmentation network, ensuring accurate attribution weights and attention maps. The parameters $\lambda$ and $\mu$  balance the three loss terms. 
Specifically:
\begin{equation}
	\begin{aligned}
	\mathcal{L}_{int}&=\frac{1}{H W}\sum_{i,j}\left(\omega_1 \odot\left(I_f-I_{i r}\right)^2+\omega_2 \odot\left(I_f-I_{v i}\right)^2\right) \\
	\mathcal{L}_{grad}&=\frac{1}{H W}\left\|\left|\nabla I_{f}\right|-\max \left(\left|\nabla I_{ir}\right|,\left|\nabla I_{vi}\right|\right)\right\|_1 \\
	\mathcal{L}_{seg}&=\frac{1}{3} [ C E\left(I_{ir}\right)+C E\left(I_{vi}\right)+C E\left(I_f\right) ],
	\end{aligned}
\end{equation}
where $\nabla$ represents the Sobel gradient operator, $|\cdot|$ denotes the absolute value, and $\max(\cdot)$ is the pixel-wise maximum operation. \
{$CE(\cdot)$ represents the cross-entropy loss for the semantic segmentation network. The $\mathcal{L}_{seg}$ term includes $I_{i r}$ and $I_{vi}$ to ensure that the segmentation network accurately computes attributions for $I_{i r}$ and $I_{vi}$ during training.}
\begin{figure*}[!]
	\centering
	\includegraphics[width=\linewidth]{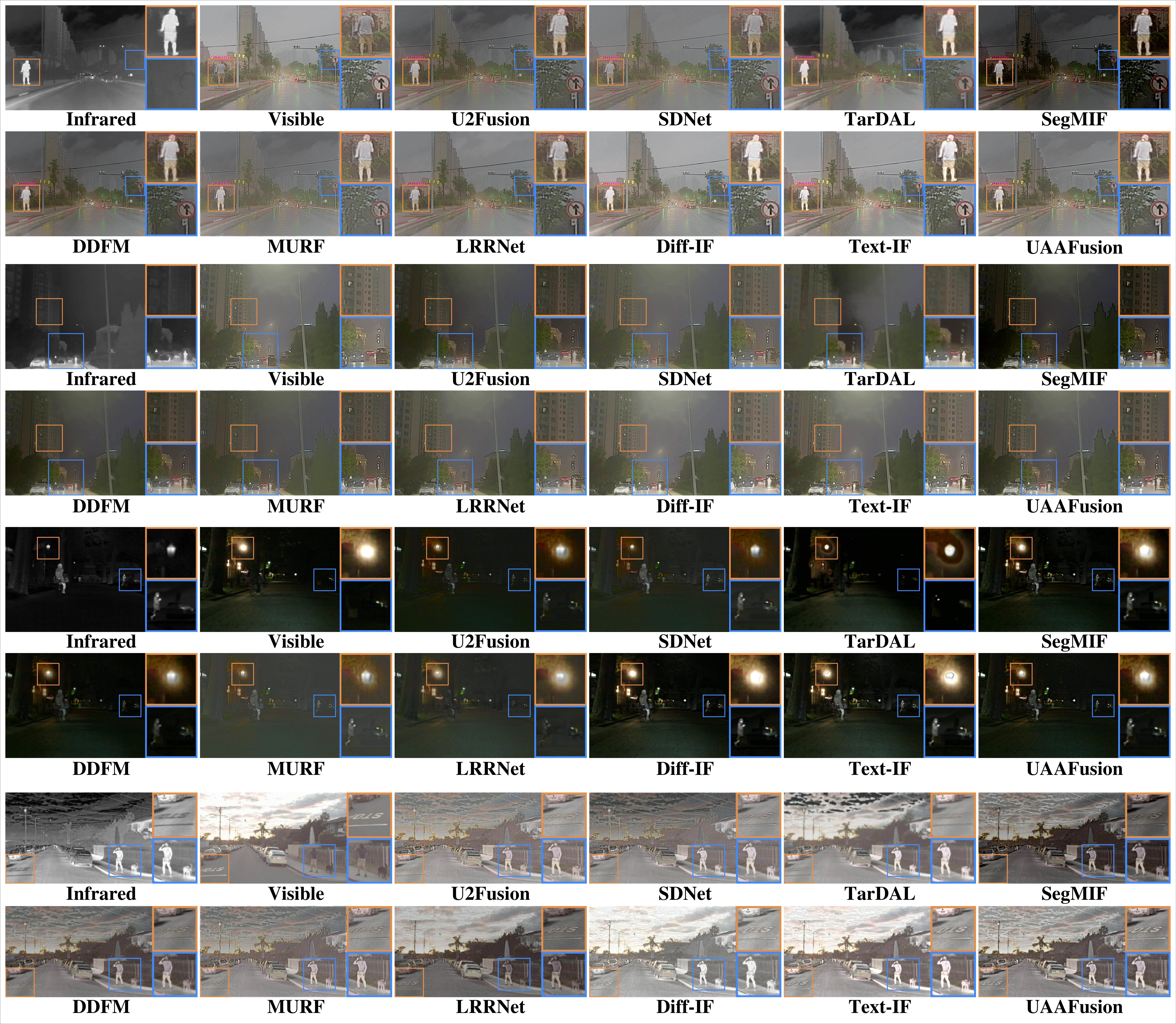}
	\caption{{Comparison of fusion results of different methods. The cases in the figure are “00122” and “00633” in FMB dataset, “00706N” in MSRS dataset and “FLIR$\_$06430” in RoadScene dataset.}}
    \vspace{-1em}
	\label{fig:Fusion}
\end{figure*}

\section{Experiment}\label{sec:4}
To assess the effectiveness of our proposed method, we conduct a series of comprehensive experiments. These experiments are meticulously designed to explore various facets of our method. Specifically, our experimental suite includes experiments on image fusion, visualization of attribution weights to understand feature contribution, comparisons with existing methods on segmentation, ablation studies to evaluate the impact of different components, and tuning of hyperparameters to optimize performance.

\subsection{Setup}
In our experiments, the hyperparameters $\lambda$ and $\mu$ are set to 1 and 0.1, respectively, based on preliminary optimization studies. We choose $K=5$ for the iterative stages, setting the initial value of the learnable step size $\rho_k$ to 0.01. We utilize 5 sampling steps $M$ to compute the attribution weights. The images from the training set are normalized to the range [0,1] and cropped into 128$\times$128 patches for processing. Training is performed in a batch size of 8 across 50 epochs, with the initial learning rate of 1e-4 being halved every 10 epochs to facilitate convergence. Adam is selected as the optimization algorithm due to its efficiency in handling sparse gradients. DeeplabV3+~\cite{DBLP:conf/eccv/ChenZPSA18}, serves as an auxiliary network for semantic segmentation to enhance the fusion process. All experiments are conducted on a PC equipped with a single NVIDIA GeForce RTX 3090 GPU, ensuring consistent computational performance.

We evaluate the fusion performance on the FMB~\cite{DBLP:journals/corr/abs-2308-02097} dataset, consisting of 1,220 training pairs and 280 test pairs.
Additionally, we assess generalization performance on 50 pairs from MSRS~\cite{DBLP:journals/inffus/TangYZJM22} and 50 pairs from RoadScene~\cite{DBLP:conf/aaai/Xu0LJG20}.
{We compare our method with several state-of-the-art methods, including U2Fusion~\cite{DBLP:journals/pami/XuMJGL22}, SDNet~\cite{DBLP:journals/ijcv/ZhangM21}, TarDAL~\cite{DBLP:conf/cvpr/LiuFHWLZL22}, SegMIF~\cite{DBLP:journals/corr/abs-2308-02097}, DDFM~\cite{zhao2023ddfm}, MURF~\cite{xu2023murf}, LRRNet~\cite{li2023lrrnet}, Diff-IF~\cite{yi2024diff}, and Text-IF~\cite{yi2024text}.
In these methods, LRRNet~\cite{li2023lrrnet} Diff-IF~\cite{yi2024diff}, and Text-IF~\cite{yi2024text} use a fixed fusion loss. DDFM~\cite{zhao2023ddfm} and Diff-IF~\cite{yi2024diff} employ diffusion models for image fusion. Like U2Fusion~\cite{DBLP:journals/pami/XuMJGL22}, SDNet~\cite{DBLP:journals/ijcv/ZhangM21}, and MURF~\cite{xu2023murf}, our method utilizes dynamic weights in the fusion loss. However, these methods do not incorporate downstream task guidance. In contrast, TarDAL~\cite{DBLP:conf/cvpr/LiuFHWLZL22} and SegMIF~\cite{DBLP:journals/corr/abs-2308-02097} integrate downstream task loss to refine their fusion processes. Our unique contribution is the use of semantic segmentation to dynamically guide the fusion process through attribution analysis, significantly enhancing the adaptability of fused images to various tasks.}

\subsection{Fusion Experiments}
We employ six metrics to evaluate the fusion results comprehensively, including entropy (EN), spatial frequency (SF), correlation coefficient (CC), visual information fidelity (VIF), $Q^{AB/F}$, and structural similarity index measure (SSIM). The methods for calculating these metrics are detailed in~\cite{DBLP:journals/inffus/MaML19}.

\begin{table}[t]	
        
	\centering
        \caption{{Quantitative results of infrared-visible image fusion. The \colorbox{firstcolor}{red} and \colorbox{secondcolor}{blue} markers represent the best and second-best values, respectively.}}
	\resizebox{\linewidth}{!}{
		\begin{tabular}{lcccccc}
			\toprule
			\multicolumn{7}{c}{\textbf{Dataset: FMB Infrared-Visible Fusion Dataset}~\cite{DBLP:journals/corr/abs-2308-02097}}                                 \\
			Methods& EN~$\uparrow$&SF~$\uparrow$&	CC~$\uparrow$&	VIF~$\uparrow$&	Qabf~$\uparrow$&	SSIM~$\uparrow$\\
			\midrule
			U2Fusion~\cite{DBLP:journals/pami/XuMJGL22} & 6.598 & 10.305 & 0.648 & 0.330 & 0.581 & \cellcolor[rgb]{ .863,  .902,  .945}{0.497} \\
			SDNet~\cite{DBLP:journals/ijcv/ZhangM21} & 6.149 & 10.859 & {0.622} & 0.319 & 0.599 & 0.485 \\
			TarDAL~\cite{DBLP:conf/cvpr/LiuFHWLZL22} & 6.630 & 6.941 & 0.506 & 0.281 & 0.287 & 0.408 \\   
			SegMIF~\cite{DBLP:journals/corr/abs-2308-02097} & \cellcolor[rgb]{ .949,  .863,  .859}{6.831} & 13.688 & 0.657 & 0.389 & 0.647 & 0.370 \\
			DDFM~\cite{zhao2023ddfm} & 6.689 & 9.018 & \cellcolor[rgb]{ .863,  .902,  .945}{0.661} & 0.338 & 0.528 & 0.455 \\
			MURF~\cite{xu2023murf} & 6.367 & 13.878 & 0.600 & 0.223 & 0.372 & 0.363 \\
			LRRNet~\cite{li2023lrrnet} & 6.283 & 10.122 & 0.641 & 0.305 & 0.553 & 0.398 \\
   {Diff-IF}~\cite{yi2024diff}& 6.632 & 13.869 & 0.583 & 0.436 & 0.660 & 0.482\\
   {Text-IF}~\cite{yi2024text}& 6.721 & \cellcolor[rgb]{ .949,  .863,  .859}{14.656} & 0.600 & \cellcolor[rgb]{ .949,  .863,  .859}{0.477} & \cellcolor[rgb]{ .863,  .902,  .945}{0.708} & 0.487\\
			UAAFusion & \cellcolor[rgb]{ .863,  .902,  .945}{6.746} & \cellcolor[rgb]{ .863,  .902,  .945}{14.243} & \cellcolor[rgb]{ .949,  .863,  .859}{0.662} & \cellcolor[rgb]{ .863,  .902,  .945}{0.438} & \cellcolor[rgb]{ .949,  .863,  .859}{0.712} & \cellcolor[rgb]{ .949,  .863,  .859}{0.499} \\
			\midrule
			\multicolumn{7}{c}{\textbf{Dataset: MSRS Infrared-Visible Fusion Dataset}~\cite{DBLP:journals/inffus/TangYZJM22}}                                 \\
			Methods& EN~$\uparrow$&SF~$\uparrow$&	CC~$\uparrow$&	VIF~$\uparrow$&	Qabf~$\uparrow$&	SSIM~$\uparrow$\\
			\midrule
			U2Fusion~\cite{DBLP:journals/pami/XuMJGL22} & 6.511 & 11.732 & \cellcolor[rgb]{ .949,  .863,  .859}{0.691} & 0.600 & 0.509 & 0.653 \\
			SDNet~\cite{DBLP:journals/ijcv/ZhangM21}     & 5.878 & 10.999 & 0.677 & 0.487 & 0.426 & 0.659 \\
			TarDAL~\cite{DBLP:conf/cvpr/LiuFHWLZL22}    & 6.428 & 6.972  & 0.492 & 0.433 & 0.222 & 0.606 \\
			SegMIF~\cite{DBLP:journals/corr/abs-2308-02097}    & 6.514 & 13.010 & 0.633 & 0.760 & 0.610 & 0.477 \\
			DDFM~\cite{zhao2023ddfm}      & 6.848 & 8.726  & 0.680 & 0.693 & 0.507 & \cellcolor[rgb]{ .863,  .902,  .945}{0.678} \\
			MURF~\cite{xu2023murf}      & 5.792 & 13.418 & 0.653 & 0.426 & 0.379 & 0.636 \\
			LRRNet~\cite{li2023lrrnet}    & 6.960 & 10.194 & 0.572 & 0.598 & 0.518 & 0.635 \\
   {Diff-IF}~\cite{yi2024diff}& \cellcolor[rgb]{ .949,  .863,  .859}{7.281} & \cellcolor[rgb]{ .863,  .902,  .945}{13.581} & 0.596 & \cellcolor[rgb]{ .949,  .863,  .859}{0.978} & \cellcolor[rgb]{ .949,  .863,  .859}{0.699} & 0.669\\
   {Text-IF}~\cite{yi2024text}& \cellcolor[rgb]{ .863,  .902,  .945}{7.197} & 13.261 & 0.595 & \cellcolor[rgb]{ .863,  .902,  .945}{0.937} & {0.674} & 0.665\\
			UAAFusion & 6.971 & \cellcolor[rgb]{ .949,  .863,  .859}{13.594} & \cellcolor[rgb]{ .863,  .902,  .945}{0.681} & 0.819 & \cellcolor[rgb]{ .863,  .902,  .945}{0.674} & \cellcolor[rgb]{ .949,  .863,  .859}{0.681} \\
			\midrule
			
			\multicolumn{7}{c}{\textbf{Dataset: RoadScene Infrared-Visible Fusion Dataset}~\cite{DBLP:conf/aaai/Xu0LJG20}}                                 \\
			Methods& EN~$\uparrow$&SF~$\uparrow$&	CC~$\uparrow$&	VIF~$\uparrow$&	Qabf~$\uparrow$&	SSIM~$\uparrow$\\
			\midrule
			U2Fusion~\cite{DBLP:journals/pami/XuMJGL22}  & \cellcolor[rgb]{ .863,  .902,  .945}{7.320} & \cellcolor[rgb]{ .863,  .902,  .945}{14.056} & 0.764 & 0.664 & 0.547 & 0.711 \\
			SDNet~\cite{DBLP:journals/ijcv/ZhangM21}     & 7.267 & 14.020 & 0.727 & 0.657 & 0.541 & 0.717 \\
			TarDAL~\cite{DBLP:conf/cvpr/LiuFHWLZL22}   & 7.220 & 10.133 & 0.726 & 0.603 & 0.427 & \cellcolor[rgb]{ .949,  .863,  .859}{0.731} \\
			SegMIF~\cite{DBLP:journals/corr/abs-2308-02097}    & 7.283 & 13.963 & 0.766 & 0.708 & 0.557 & 0.598 \\
			DDFM~\cite{zhao2023ddfm}      & 7.310 & 10.451 & \cellcolor[rgb]{ .949,  .863,  .859}{0.786} & 0.685 & 0.512 & 0.715 \\
			MURF~\cite{xu2023murf}      & 6.882 & 13.056 & 0.753 & 0.553 & 0.459 & 0.715 \\
			LRRNet~\cite{li2023lrrnet}    & 7.119 & 10.795 & 0.775 & 0.496 & 0.356 & 0.619 \\
        {Diff-IF}~\cite{yi2024diff}& 7.264 & 13.704 & 0.749 & \cellcolor[rgb]{ .863,  .902,  .945}{0.722} & \cellcolor[rgb]{ .863,  .902,  .945}{0.582} & 0.712\\
        {Text-IF}~\cite{yi2024text}& 7.057 & 11.689 & 0.716 & 0.664 & 0.500 & 0.706\\
   
			UAAFusion & \cellcolor[rgb]{ .949,  .863,  .859}{7.380} & \cellcolor[rgb]{ .949,  .863,  .859}{15.645} & \cellcolor[rgb]{ .863,  .902,  .945}{0.774} & \cellcolor[rgb]{ .949,  .863,  .859}{0.729} & \cellcolor[rgb]{ .949,  .863,  .859}{0.598} & \cellcolor[rgb]{ .863,  .902,  .945}{0.713} \\
			\bottomrule
		\end{tabular}
	}
     \vspace{-1em}
 \label{tab:Fusion}
\end{table}

\begin{figure}[!]
	\centering
	\includegraphics[width=\linewidth]{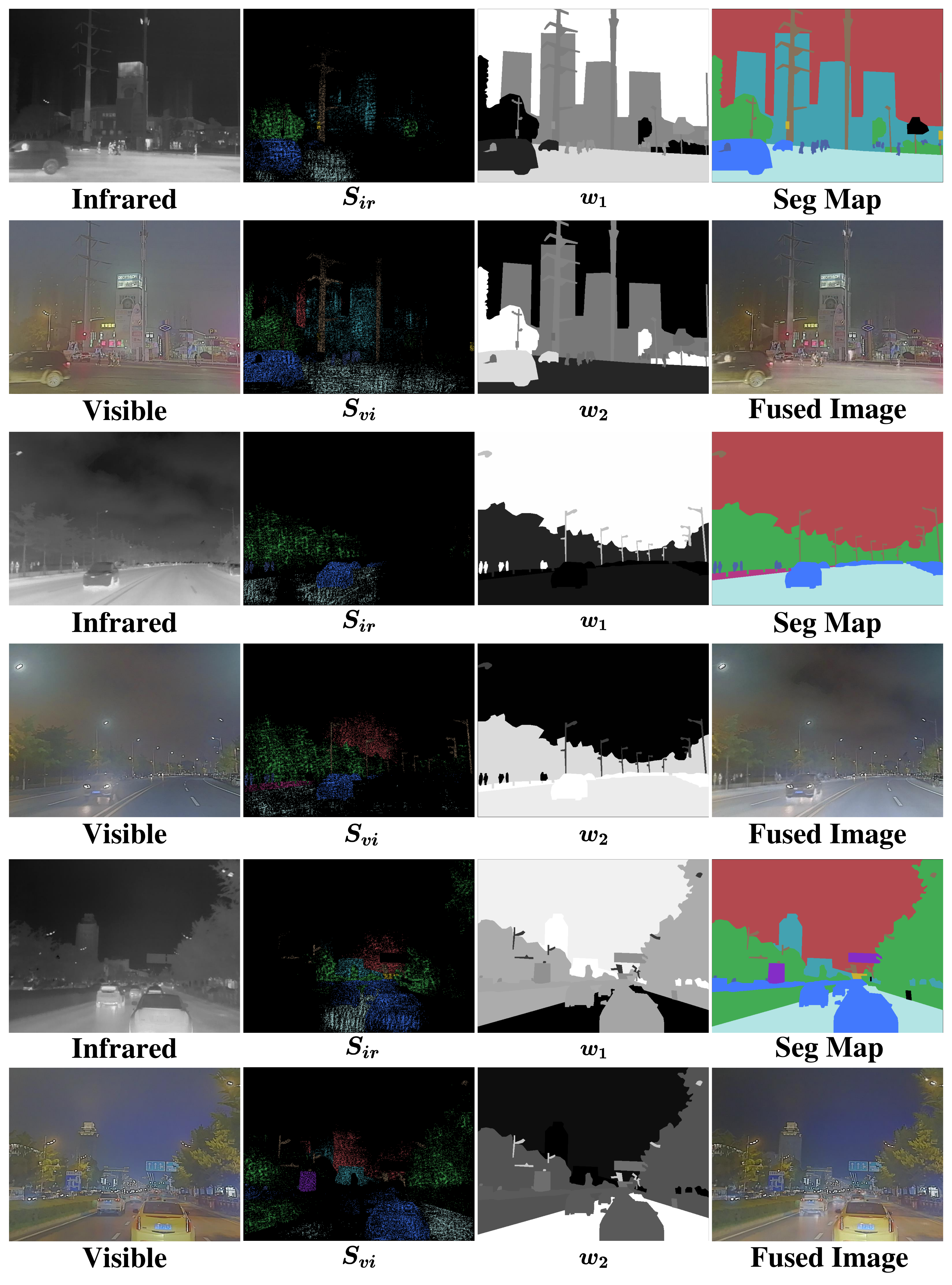}
	\caption{{Visualization of attribution weights for "00524", “00703” and “00920” in FMB dataset.}}
	\label{fig:weight}
    \vspace{-1em}
\end{figure}

\begin{figure*}[!]
	\centering
	\includegraphics[width=\linewidth]{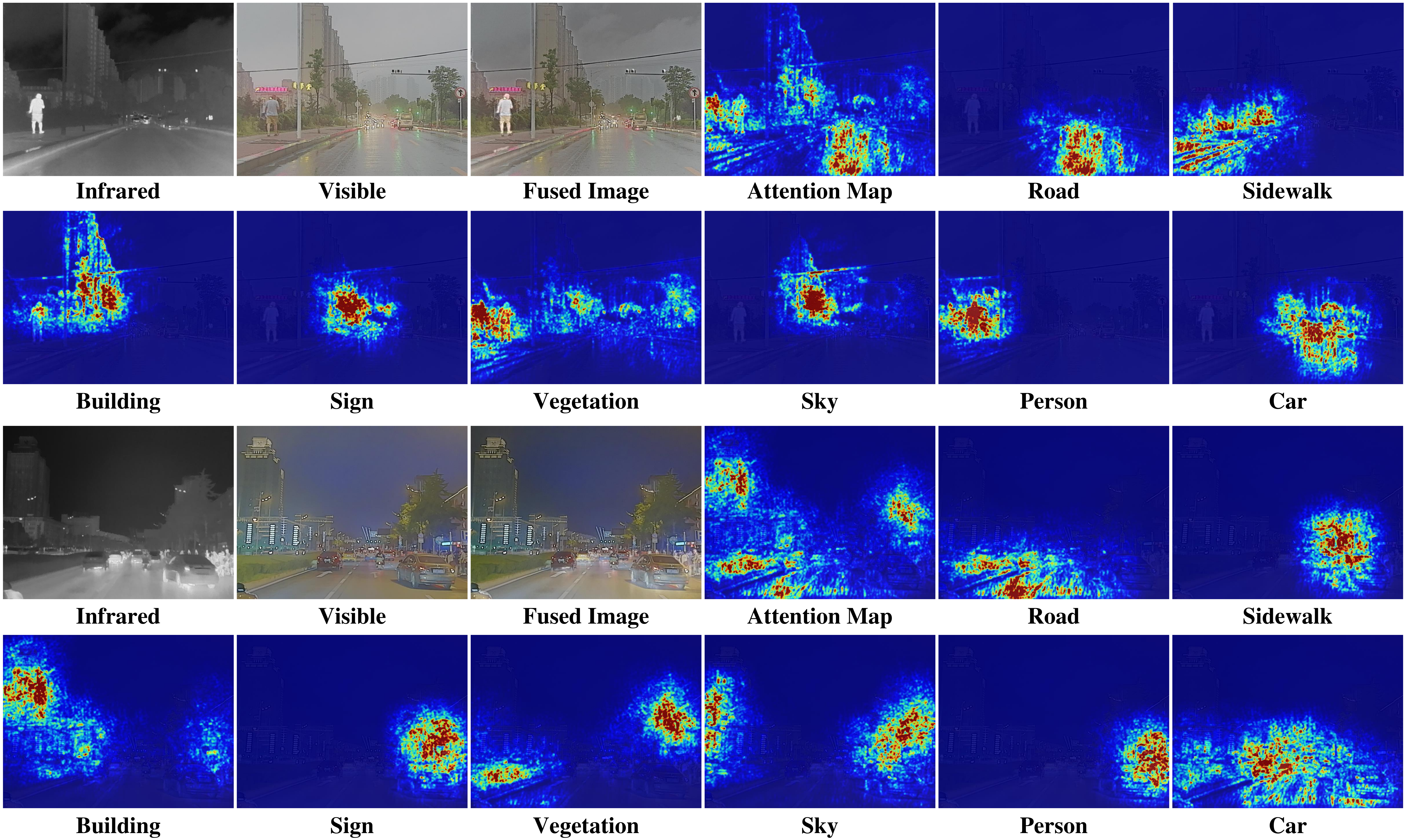}
	\caption{{Visualization of the attribution attention map for the total attention and some semantic classes. The cases in the figure are “00122” and “00997” in FMB dataset.}}
	\label{fig:att}
\end{figure*}

\subsubsection{Qualitative comparison}
The fusion performance of our method on the FMB, MSRS, and RoadScene datasets is illustrated in Fig.~\ref{fig:Fusion}.
The fusion of infrared and visible images enhances scene understanding.
In the first example, depicting a daytime street scene, our method more effectively preserves the thermal radiation information of people compared to other methods and maintains the texture details of traffic signs.
The second example depicts scenes affected by inadequate night lighting and halo effects, our method continues to provide clear visual results, unaffected by these environmental factors.
The last two examples demonstrate the superior generalization performance of UAAFusion.
Compared to other methods, UAAFusion excels at retaining critical edge and texture details, such as signs and buildings, enhancing the visual readability of images.
Furthermore, our method effectively maintains the original contrast and color fidelity of the images.
The images produced by UAAFusion display enhanced color richness and naturalness, closely resembling the true scenes under natural observation conditions.

\subsubsection{Quantitative comparison}
Our model is trained and tested on the FMB dataset and subsequently generalized to 50 images from MSRS and 50 images from RoadScene without any fine-tuning. Quantitative comparisons are presented in Tab.~\ref{tab:Fusion}. UAAFusion consistently outperforms competing methods across all three datasets, achieving high rankings on multiple evaluation metrics, thereby underscoring its stability and effectiveness in various scenarios. Consistently high entropy (EN) values indicate that images produced by UAAFusion are rich in information. The spatial frequency (SF) and $Q^{AB/F}$ scores demonstrate the ability of our method to produce clear images while retaining detailed information. The highest visual information fidelity (VIF) scores highlight our method's robust capability in visual information fidelity, suggesting excellent alignment with the human visual system. The correlation coefficient (CC) and structural similarity index measure (SSIM) scores affirm the similarity and correlation between the fused and original images, further proving our method's capability to preserve information.

\subsection{Visualization of the Attribution Weight}
Our image fusion framework incorporates attribution weights derived from attribution analysis. Fig.~\ref{fig:weight} displays several activation maps showing attribution weights and their computed results. The first example, an evening streetscape, shows that under such lighting conditions, the visible image retains clearer vehicle features. Consequently, the vehicle area in the visible image is assigned greater weight. Conversely, the thermal radiation information from people in the infrared image, crucial for segmentation, receives greater weight. In the second example, the most notable feature is the streetlight. In the visible image, floodlighting renders the streetlight features nearly indistinguishable, whereas the infrared image retains its complete structure. As a result, under the influence of attribution weights, the fused image displays the streetlight's structure. Contrary to the second example, the third example shows a scene where the visible image retains a better structure of the streetlight. Therefore, the weight distribution in the streetlight area is the opposite of that in the second example. Additionally, while buildings in the distance have blurred boundaries in the visible image, the infrared image's clear boundaries facilitate segmentation, resulting in distinct architectural outlines in the fused image.

\subsection{Visualization of the Attribution Attention Map}
In Fig.~\ref{fig:att},  we present the effects of our attribution attention mechanism in various scenarios, highlighting the most representative category-specific attention. The visual results demonstrate that our auxiliary semantic segmentation network effectively identifies and emphasizes the image areas critical for segmentation through attribution analysis. This enhances the efficiency of our fusion network's focus. Although discrepancies exist between the attention areas in Fig.~\ref{fig:att} and the actual targets, these areas significantly enhance the downstream task networks' discrimination capabilities. They do more than just pinpoint target locations.

This attention mechanism focuses on areas that directly impact task performance. This focus optimizes the model's processing and enhances both computational efficiency and task performance. For example, in road and pedestrian detection scenarios, the mechanism highlights critical areas like roads and sidewalks, essential for real-time decision-making in autonomous vehicles. Furthermore, applying this mechanism across various scenes effectively distinguishes elements like the sky, buildings, and vehicles. This approach not only minimizes attention dispersion to less critical areas but also deepens the understanding of regions vital for outcome determination.

The attribution attention mechanism precisely controls the model's focus, substantially enhancing its performance and efficiency in real-world applications. This method, based on attribution analysis, allows the model to focus more on key information that aids in task completion within complex environments, significantly improving the accuracy and reliability of decision-making.

\subsection{Downstream Applications}
\begin{figure*}[!]
	\centering
	\includegraphics[width=\linewidth]{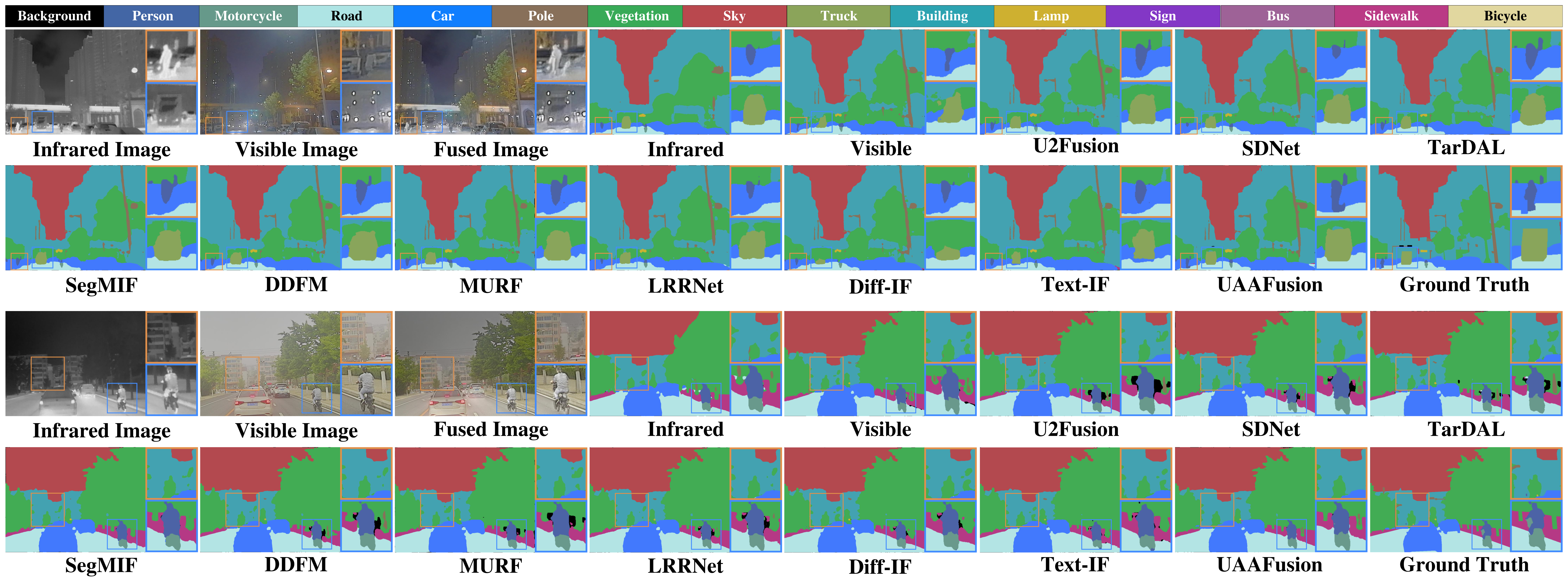}
	\caption{{Visual comparison of semantic segmentation task for ``01046'' and “01439” in FMB dataset.}}
	\label{fig:Seg}
 \vspace{-1em}
\end{figure*}

\begin{table*}[!htbp]
	\centering
	\caption{Quantitative semantic segmentation results of different methods on the FMB dataset. The \colorbox{firstcolor}{red} and \colorbox{secondcolor}{blue} markers represent the best and second-best values, respectively.}
	\label{tab:Seg}
	\resizebox{\textwidth}{!}{
		\begin{tabular}{lcccccccccccccccccc}
			\toprule
			\rowcolor{white}
			& \multicolumn{2}{c}{Car} & \multicolumn{2}{c}{Person} & \multicolumn{2}{c}{Truck} & \multicolumn{2}{c}{Lamp} & \multicolumn{2}{c}{Sign} & \multicolumn{2}{c}{Building} & \multicolumn{2}{c}{Vegetation} & \multicolumn{2}{c}{Pole} &  &  \\
			\cmidrule(lr){2-3} \cmidrule(lr){4-5} \cmidrule(lr){6-7} \cmidrule(lr){8-9} \cmidrule(lr){10-11} \cmidrule(lr){12-13} \cmidrule(lr){14-15} \cmidrule(lr){16-17}
			Methods & Acc & IoU & Acc & IoU & Acc & IoU & Acc & IoU & Acc & IoU & Acc & IoU & Acc & IoU & Acc & IoU & mAcc & mIoU \\
			\midrule
                Infrared & 92.64 & 76.02                     & 74.24 & 66.45 & 19.96 & 19.34 & 23.23 & 22.07 & 66.07 & 60.55 & 89.89 & 80.62 & 90.74 & 81.25 & 41.03 & 30.91 & 62.23 & 54.65 \\
            
			Visible   & 93.05 & 80.56                     & 64.79 & 56.26 & 30.87 & 29.26 & 31.50 & 29.48 & 82.31 & 70.52 & 90.19 & 80.61 & 92.58 & 84.70 & 58.90 & 45.29 & 68.02 & 59.59 \\
			U2Fusion~\cite{DBLP:journals/pami/XuMJGL22}  & 93.40 & 80.71                     & 74.35 & 66.70 & 30.41 & 28.77 & 34.47 & 31.18 & 84.82 & \cellcolor[rgb]{ .863,  .902,  .945}{73.15} & 91.27 & \cellcolor[rgb]{ .863,  .902,  .945}{82.63} & \cellcolor[rgb]{ .863,  .902,  .945}{93.59} & \cellcolor[rgb]{ .863,  .902,  .945}{86.36} & 60.77 & 47.60 & 70.39 & 62.14 \\
			SDNet~\cite{DBLP:journals/ijcv/ZhangM21}     & 93.51 & 80.75                     & 72.17 & 64.95 & 37.71 & 35.93 & 32.95 & 30.71 & 83.33 & 71.36 & 91.59 & 81.78 & 93.12 & 86.14 & 60.22 & 46.68 & 70.57 & 62.29 \\
			TarDAL~\cite{DBLP:conf/cvpr/LiuFHWLZL22}    & 93.21 & 80.23                     & 71.19 & 64.91 & 17.69 & 17.13 & 29.92 & 28.24 & \cellcolor[rgb]{ .863,  .902,  .945}{84.98} & 72.38 & 90.25 & 80.81 & 93.06 & 85.48 & 59.14 & 45.65 & 67.43 & 59.35 \\
			SegMIF~\cite{DBLP:journals/corr/abs-2308-02097}    & 93.16 & 79.80                     & \cellcolor[rgb]{ .949,  .863,  .859}{75.73} & \cellcolor[rgb]{ .863,  .902,  .945}{67.23} & \cellcolor[rgb]{ .863,  .902,  .945}{43.62} & \cellcolor[rgb]{ .949,  .863,  .859}{40.50} & 33.53 & 30.15 & 84.27 & 72.55 & 91.11 & 82.19 & 93.10 & 85.68 & 59.50 & 45.87 & \cellcolor[rgb]{ .863,  .902,  .945}{71.75} & \cellcolor[rgb]{ .863,  .902,  .945}{63.00}  \\
			DDFM~\cite{zhao2023ddfm}     & 93.64 & \cellcolor[rgb]{ .863,  .902,  .945}{81.09} & 72.92 & 65.77 & 30.33 & 28.74 & \cellcolor[rgb]{ .949,  .863,  .859}{40.83} & \cellcolor[rgb]{ .949,  .863,  .859}{36.83} & 83.95 & 72.69 & 91.90 & 82.50 & 93.49 & \cellcolor[rgb]{ .949,  .863,  .859}{86.39} & 60.88 & \cellcolor[rgb]{ .863,  .902,  .945}{47.95} & 70.99 & 62.75 \\
			MURF~\cite{xu2023murf}  & {93.69} & 80.10                     & 72.60 & 65.83 & 23.31 & 22.41 & 31.50 & 29.28 & 82.48 & 69.82 & \cellcolor[rgb]{ .949,  .863,  .859}{92.55} & 82.27 & 93.10 & 86.30 & \cellcolor[rgb]{ .863,  .902,  .945}{61.21} & 47.10 & 68.80 & 60.39    \\
			LRRNet~\cite{li2023lrrnet}    & 93.22 & \cellcolor[rgb]{ .949,  .863,  .859}{81.45}                     & 74.14 & 66.16 & 39.00 & 36.74 & 34.95 & 32.51 & 82.58 & 71.48 & 91.67 & 82.00 & 92.95 & 85.75 & 60.55 & 47.06 & 71.13 & 62.90 \\
            {Diff-IF}~\cite{yi2024diff} & 93.72 & 76.96 & 71.44 & 65.03 & 8.24  & 8.09  & 29.42 & 27.20 & 82.18 & 70.53 & 89.15 & 80.62 & {93.52} & 84.55 & 57.85 & 45.51 & 65.69 & 57.31 \\
            {Text-IF}~\cite{yi2024text} & \cellcolor[rgb]{ .863,  .902,  .945}{93.84} & 79.68 & \cellcolor[rgb]{ .863,  .902,  .945}{74.46} & 66.77 & 39.89 & 34.82 & 35.50 & 32.98 & 83.56 & 72.24 & 90.38 & 82.44 & {93.13} & 84.96 & 58.54 & 46.31 & 71.16 & 62.52 \\
			UAAFusion & \cellcolor[rgb]{ .949,  .863,  .859}{94.24} & 80.89                     & {74.36} & \cellcolor[rgb]{ .949,  .863,  .859}{67.60} & \cellcolor[rgb]{ .949,  .863,  .859}{44.39} & \cellcolor[rgb]{ .863,  .902,  .945}{39.56} & \cellcolor[rgb]{ .863,  .902,  .945}{35.77} & \cellcolor[rgb]{ .863,  .902,  .945}{33.72} & \cellcolor[rgb]{ .949,  .863,  .859}{85.75} & \cellcolor[rgb]{ .949,  .863,  .859}{75.44} & \cellcolor[rgb]{ .863,  .902,  .945}{92.09} & \cellcolor[rgb]{ .949,  .863,  .859}{83.76} & \cellcolor[rgb]{ .949,  .863,  .859}{93.60} & 86.13 & \cellcolor[rgb]{ .949,  .863,  .859}{62.10} & \cellcolor[rgb]{ .949,  .863,  .859}{49.34} & \cellcolor[rgb]{ .949,  .863,  .859}{72.79} & \cellcolor[rgb]{ .949,  .863,  .859}{64.55}\\
			
			\bottomrule
		\end{tabular}
	}
   \vspace{-1em}
\end{table*}
Our fusion method not only improves performance but also demonstrates greater applicability in semantic segmentation. Semantic segmentation experiments are conducted on the FMB dataset~\cite{DBLP:journals/corr/abs-2308-02097}, which includes 1500 image pairs (1220 for training, 280 for testing), encompassing 14 pixel-level categories including unlabeled, road, sidewalk, building, lamp, sign, vegetation, sky, person, car, truck, bus, motorcycle, bicycle, and pole. To ensure a fair comparison, we employ SegFormer~\cite{xie2021segformer} as the benchmark network, retraining it with the cross-entropy loss function to ensure a fair comparison across all fusion methods. The segmentation results of different fusion methods are evaluated using intersection over union (IoU) and accuracy (Acc). We set the batch size to 8. The AdamW optimizer, with a momentum of 0.9, is employed for optimization. Training extends over 300 epochs, starting with a learning rate of 1e-4 and gradually decreasing to 1e-6. Comparative results are shown in Tab.~\ref{tab:Seg} and Fig.~\ref{fig:Seg}.

Experimental results indicate that UAAFusion achieves higher accuracy and IoU across several categories, including car, building, and vegetation. In the car category, UAAFusion achieved an accuracy of 94.24\% and an IoU of 80.89\%, showcasing its strong capability to distinguish vehicles—a critical skill in urban scenarios. UAAFusion's accuracy in the building category reaches 92.09\% with an IoU of 83.76\%, highlighting its effectiveness in distinguishing building structures essential for urban planning and navigation. For vegetation, UAAFusion impressively records an accuracy of 93.60\% and an IoU of 86.13\%, excelling in natural environment segmentation tasks. Individual infrared or visible images generally underperform compared to the fusion method. This suggests that fusion provides more information, aiding the model in more accurately understanding and parsing scenes. Nonetheless, compared to other methods, UAAFusion demonstrates superior results on most evaluation metrics, particularly in handling complex and diverse scenes. The visual results reveal that UAAFusion's segmentation is closer to the ground truth, especially in handling fine details accurately. For example, UAAFusion excels in identifying closely spaced or partially obscured vehicles and persons, a crucial advantage in practical applications. UAAFusion's excellent performance in semantic segmentation tasks confirms its strong capability to handle multi-source data, thereby significantly enhancing downstream task performance. Furthermore, its capability to preserve outstanding detail and deliver high-precision output is crucial for improving the reliability and efficiency of automated systems. This makes UAAFusion particularly valuable in applications such as autonomous driving, urban monitoring, and environmental surveillance.
\subsection{Ablation Studies}
We conduct a series of ablation studies to explore the contributions of the modules in our proposed network. We compared the fusion effects across all ablation settings under various conditions. The results of these investigations are presented in Tab.~\ref{tab:Ablation}. Initially, we conduct an ablation study on attributed attention Exp.~\uppercase\expandafter{\romannumeral1} by removing the proposed attributed attention. In Exp.~\uppercase\expandafter{\romannumeral2}, we modify the Integrated Gradient (IG) $\sum_{j=1}^k \frac{\partial D S\left(I_{f}^{(j)}\right)}{\partial (I_{f}^{(j)})_{i}} \cdot\left(I_{f}^{(j)}-I_{f}^{(j-1)}\right)_{i}$ used in the attention map computation to a directly solved gradient (Grad): $\partial D S\left(I_{f}^{(k)}\right)/{\partial (I_{f}^{(k)})_{i}}$. We then explored the impact of different loss functions. 
We remove the intensity loss in Exp.~\uppercase\expandafter{\romannumeral3} and the gradient loss in Exp.~\uppercase\expandafter{\romannumeral4} to verify the validity of the loss functions. Finally, we explored the contributions of the memory augmentation module. In Exp.~\uppercase\expandafter{\romannumeral5}, we remove the short-term memory, followed by the removal of long-term memory in Exp.~\uppercase\expandafter{\romannumeral6}. Exp.~\uppercase\expandafter{\romannumeral7} investigates the scenario where both short-term and long-term memory augmentations are absent.
{In Exp.~\uppercase\expandafter{\romannumeral8}, we modify the segmentation loss by not calculating the cross-entropy with respect to $I_{ir}$ and $I_{vi}$.}

\begin{table}[t]
	\centering
	\caption{{Quantitative results of ablation studies for image fusion experiments. The \colorbox{firstcolor}{red} marker represents the best value, respectively.}}
	\label{tab:Ablation}
	\resizebox{\linewidth}{!}{
	\begin{tabular}{cccccccc}
		\toprule
		
		\multicolumn{8}{c}{\textbf{Ablation Studies}} \\
		&Exp & EN~$\uparrow$&SF~$\uparrow$&	CC~$\uparrow$&	VIF~$\uparrow$&	Qabf~$\uparrow$&	SSIM~$\uparrow$ \\ \midrule
		\uppercase\expandafter{\romannumeral1} &	w/o $att$ &6.703 & 13.262 & 0.611 & 0.419 & 0.699 & 0.494  \\
		\uppercase\expandafter{\romannumeral2} &IG $\rightarrow$ Grad & 6.722 & 13.926 & 0.637 & 0.427 & 0.692 & 0.491  \\
		\uppercase\expandafter{\romannumeral3} &w/o $L_{int}$ & 6.692 & 13.511 & 0.629 & 0.394 & 0.705 & 0.458  \\
		\uppercase\expandafter{\romannumeral4}&w/o $L_{grad}$ & 6.697 & 13.197 & 0.645 & 0.385 & 0.663 & 0.463  \\
		\uppercase\expandafter{\romannumeral5}&w/o $m_{s}$ & 6.713 & 13.206 & 0.633 & 0.403 & 0.674 & 0.472 \\
		\uppercase\expandafter{\romannumeral6}&w/o $m_{l}$ & 6.719 & 13.228 & 0.634 & 0.410  & 0.679 & 0.479  \\
		\uppercase\expandafter{\romannumeral7}&w/o $m_{s} + m_{l}$ & 6.711 & 13.146 & 0.623 & 0.383 & 0.670  & 0.469  \\ 
  {\uppercase\expandafter{\romannumeral8}}& {$\mathcal{L}_{seg}\!=\!CE\left(I_f\right)$}  & 6.682 & 13.349 & 0.622 & 0.416 & 0.638  & 0.466  \\ \midrule
  
		&Ours & \cellcolor[rgb]{ .949,  .863,  .859}{6.746} & \cellcolor[rgb]{ .949,  .863,  .859}{14.243} & \cellcolor[rgb]{ .949,  .863,  .859}{0.662} & \cellcolor[rgb]{ .949,  .863,  .859}{0.438} & \cellcolor[rgb]{ .949,  .863,  .859}{0.712} & \cellcolor[rgb]{ .949,  .863,  .859}{0.499}  \\ \bottomrule
	\end{tabular}}%
        \vspace{-1em}
\end{table}%

\begin{table}[t]
	\centering
	\caption{{Investigation of parameters on the FMB dataset. The \colorbox{firstcolor}{red} and \colorbox{secondcolor}{blue} markers represent the best and second-best values.}}
	\label{tab:para}

		\resizebox{0.85\linewidth}{!}{
		\begin{tabular}{ccccccc}
			\toprule
			\multicolumn{7}{c}{\textbf{Stage number}} \\
			$K$ & EN~$\uparrow$&SF~$\uparrow$&	CC~$\uparrow$&	VIF~$\uparrow$&	Qabf~$\uparrow$&	SSIM~$\uparrow$ \\ \midrule
			2  &6.281 & 13.946 & 0.617 & 0.375 & 0.694 & 0.457 \\ 
			3  &6.718 & 14.198 & 0.655 & 0.427 & 0.702 & 0.462 \\ 
			4  &6.739 & 14.217 & 0.657 & 0.433 & 0.708 & 0.486 \\ 
			5  &\cellcolor[rgb]{ .863,  .902,  .945}{6.746} & \cellcolor[rgb]{ .863,  .902,  .945}{14.243} & \cellcolor[rgb]{ .949,  .863,  .859}{0.662} & \cellcolor[rgb]{ .949,  .863,  .859}{0.438} & \cellcolor[rgb]{ .949,  .863,  .859}{0.712} & \cellcolor[rgb]{ .949,  .863,  .859}{0.499} \\ 
			6  &6.745 & \cellcolor[rgb]{ .949,  .863,  .859}{14.256} & \cellcolor[rgb]{ .863,  .902,  .945}{0.661} & 0.436 & \cellcolor[rgb]{ .863,  .902,  .945}{0.711} & 0.497\\ 
			7  &\cellcolor[rgb]{ .949,  .863,  .859}{6.748} & 14.241 & 0.661 & \cellcolor[rgb]{ .863,  .902,  .945}{0.436} & 0.708 & \cellcolor[rgb]{ .863,  .902,  .945}{0.498}\\ 
			8  &6.746 & 14.237 & 0.662 & 0.429 & 0.710  & 0.498\\ \midrule 
			\multicolumn{7}{c}{{\textbf{Approximation Steps in Eq. (12)}}} \\
   {Num} & {EN~$\uparrow$}&{SF~$\uparrow$}&	{CC~$\uparrow$}&	{VIF~$\uparrow$}&	{Qabf~$\uparrow$}&	{SSIM~$\uparrow$ }\\ \midrule
           $k$  & 6.746 & \cellcolor[rgb]{ .863,  .902,  .945}{14.243} & \cellcolor[rgb]{ .863,  .902,  .945}{0.662} & 0.438 & \cellcolor[rgb]{ .863,  .902,  .945}{0.712} & 0.499 \\
            $2k$ & \cellcolor[rgb]{ .863,  .902,  .945}{6.748} & 14.239 & 0.660 & \cellcolor[rgb]{ .949,  .863,  .859}{0.440} & 0.711 & \cellcolor[rgb]{ .949,  .863,  .859}{0.500} \\
            $3k$ & 6.746 & 14.242 & 0.661 & \cellcolor[rgb]{ .863,  .902,  .945}{0.439} & 0.712 & \cellcolor[rgb]{ .863,  .902,  .945}{0.499} \\
            $4k$ & \cellcolor[rgb]{ .949,  .863,  .859}{6.749} & 14.214 & 0.662 & 0.435 & 0.708 & 0.497 \\
            $5k$ & 6.747 & \cellcolor[rgb]{ .949,  .863,  .859}{14.244} & \cellcolor[rgb]{ .949,  .863,  .859}{0.663} & 0.438 & \cellcolor[rgb]{ .949,  .863,  .859}{0.713} & 0.498 \\
   
			\bottomrule

	\end{tabular}}%
 \vspace{-1em}
\end{table}%

The results indicate that attribution attention directs the network's focus toward regions relevant to segmentation, thereby enabling the generation of higher-quality images. Changing the computation method of the attention map leads to a slight decrease in image quality. This is because directly computing gradients does not accurately measure feature contributions. The ablation study of the loss functions demonstrates that both intensity loss and gradient loss are crucial for ensuring the fidelity of fused images.
{Experiments on segmentation loss also demonstrate that its components guarantee fusion performance.} Eliminating the memory augmentation module degrades all performance metrics. This module plays a critical role in reducing information loss during feature channel switching and enhancing interaction between different processing stages.

\subsection{Parameter Investigation}
In this section, we assess how the number of stages ($K$) and the approximation steps per stage affect network performance, as specified in Eq.~(\ref{equ8}). Tab.~\ref{tab:para} presents the quantitative outcomes for various settings of $K$. The model's performance improves with an increase in the number of stages up to $K=5$, benefiting from its iterative unfolding and robust feature extraction capabilities. However, when $K$ exceeds 5, a slight decline in performance is observed, possibly due to overfitting. Therefore, we selected $K=5$ as the optimal configuration to achieve a balance between performance and complexity.

{Moreover, our tests on the number of approximation steps reveal that setting the sampling rate equal to the stage number, $k$, adequately simulates the image's transformation process and produces an effective attribution map. Further increasing the approximation steps does not substantially enhance performance and leads to a corresponding rise in computational demands. Hence, we opted to retain our strategy of conducting $k$ approximations for each stage when calculating the attribution map, optimizing both accuracy and efficiency.}

\section{Conclusion}
\label{sec:5}
We introduce an attribution-based fusion algorithm utilizing a semantic segmentation network. This network boosts the algorithm's effectiveness by aggregating critical information from the source images and prioritizing areas of high attribution importance. In this paper, we employ attribution analysis to evaluate the contribution of features in the source images to semantic segmentation. This evaluation helps establish a targeted optimization objective. The structure of the fusion network, derived from this optimization objective, features layers constructed using an iterative optimization process. This architecture ensures better alignment between the network's capabilities, algorithmic goals, and attribution attention. Furthermore, attribution analysis within each network layer pinpoints critical areas in fused images that significantly enhance semantic segmentation. Extensive experiments confirm that our algorithm consistently delivers high-quality fused images and adapts effectively to segmentation tasks. {Despite its effectiveness, the proposed approach incurs a longer runtime than fusion methods trained with a fixed loss function. This increased runtime stems from the complex computations required for the attribution analysis in semantic segmentation. In future work, we aim to: 1) enhance the computational efficiency and reduce the processing time of the attribution analysis, and explore faster, more effective fusion guidance methods via semantic segmentation; 2) apply the concepts of this study to a wider array of downstream tasks beyond semantic segmentation.}

\bibliographystyle{IEEEtran}
\bibliography{refer}

\begin{thebibliography}{100}
\providecommand{\url}[1]{#1}
\csname url@samestyle\endcsname
\providecommand{\newblock}{\relax}
\providecommand{\bibinfo}[2]{#2}
\providecommand{\BIBentrySTDinterwordspacing}{\spaceskip=0pt\relax}
\providecommand{\BIBentryALTinterwordstretchfactor}{4}
\providecommand{\BIBentryALTinterwordspacing}{\spaceskip=\fontdimen2\font plus
\BIBentryALTinterwordstretchfactor\fontdimen3\font minus \fontdimen4\font\relax}
\providecommand{\BIBforeignlanguage}[2]{{%
\expandafter\ifx\csname l@#1\endcsname\relax
\typeout{** WARNING: IEEEtran.bst: No hyphenation pattern has been}%
\typeout{** loaded for the language `#1'. Using the pattern for}%
\typeout{** the default language instead.}%
\else
\language=\csname l@#1\endcsname
\fi
#2}}
\providecommand{\BIBdecl}{\relax}
\BIBdecl

\bibitem{tang2023datfuse}
W.~Tang, F.~He, Y.~Liu, Y.~Duan, and T.~Si, ``Datfuse: Infrared and visible image fusion via dual attention transformer,'' \emph{IEEE Transactions on Circuits and Systems for Video Technology}, vol.~33, no.~7, pp. 3159--3172, 2023.

\bibitem{sun2022drone}
Y.~Sun, B.~Cao, P.~Zhu, and Q.~Hu, ``Drone-based rgb-infrared cross-modality vehicle detection via uncertainty-aware learning,'' \emph{IEEE Transactions on Circuits and Systems for Video Technology}, vol.~32, no.~10, pp. 6700--6713, 2022.

\bibitem{park2023cross}
S.~Park, A.~G. Vien, and C.~Lee, ``Cross-modal transformers for infrared and visible image fusion,'' \emph{IEEE Transactions on Circuits and Systems for Video Technology}, vol.~34, no.~2, pp. 770--785, 2023.

\bibitem{liu2021learning}
J.~Liu, X.~Fan, J.~Jiang, R.~Liu, and Z.~Luo, ``Learning a deep multi-scale feature ensemble and an edge-attention guidance for image fusion,'' \emph{IEEE Transactions on Circuits and Systems for Video Technology}, vol.~32, no.~1, pp. 105--119, 2021.

\bibitem{zhao2023tufusion}
Y.~Zhao, Q.~Zheng, P.~Zhu, X.~Zhang, and W.~Ma, ``Tufusion: A transformer-based universal fusion algorithm for multimodal images,'' \emph{IEEE Transactions on Circuits and Systems for Video Technology}, 2023.

\bibitem{li2023ccafusion}
X.~Li, Y.~Li, H.~Chen, Y.~Peng, and P.~Pan, ``Ccafusion: cross-modal coordinate attention network for infrared and visible image fusion,'' \emph{IEEE Transactions on Circuits and Systems for Video Technology}, 2023.

\bibitem{DBLP:journals/corr/abs-2211-14461}
Z.~Zhao, H.~Bai, J.~Zhang, Y.~Zhang, S.~Xu, Z.~Lin, R.~Timofte, and L.~Van~Gool, ``Cddfuse: Correlation-driven dual-branch feature decomposition for multi-modality image fusion,'' in \emph{Proceedings of the IEEE/CVF Conference on Computer Vision and Pattern Recognition (CVPR)}.\hskip 1em plus 0.5em minus 0.4em\relax {Computer Vision Foundation / IEEE}, June 2023, pp. 5906--5916.

\bibitem{yang2021infrared}
Y.~Yang, J.~Liu, S.~Huang, W.~Wan, W.~Wen, and J.~Guan, ``Infrared and visible image fusion via texture conditional generative adversarial network,'' \emph{IEEE Transactions on Circuits and Systems for Video Technology}, vol.~31, no.~12, pp. 4771--4783, 2021.

\bibitem{liu2024task}
R.~Liu, Z.~Liu, J.~Liu, X.~Fan, and Z.~Luo, ``A task-guided, implicitly-searched and metainitialized deep model for image fusion,'' \emph{IEEE Transactions on Pattern Analysis and Machine Intelligence}, 2024.

\bibitem{liu2023paif}
Z.~Liu, J.~Liu, B.~Zhang, L.~Ma, X.~Fan, and R.~Liu, ``Paif: Perception-aware infrared-visible image fusion for attack-tolerant semantic segmentation,'' \emph{ACM MM}, 2023.

\bibitem{DBLP:journals/inffus/MaML19}
J.~Ma, Y.~Ma, and C.~Li, ``Infrared and visible image fusion methods and applications: {A} survey,'' \emph{Information Fusion}, vol.~45, pp. 153--178, 2019.

\bibitem{DBLP:journals/tcsv/LiuCR20}
S.~Liu, J.~Chen, and S.~Rahardja, ``A new multi-focus image fusion algorithm and its efficient implementation,'' \emph{IEEE Transactions on Circuits and Systems for Video Technology}, vol.~30, no.~5, pp. 1374--1384, 2020.

\bibitem{DBLP:conf/aaai/JingLDWDSW20}
Y.~Jing, X.~Liu, Y.~Ding, X.~Wang, E.~Ding, M.~Song, and S.~Wen, ``Dynamic instance normalization for arbitrary style transfer,'' in \emph{Proceedings of the AAAI conference on artificial intelligence (AAAI)}, 2020, pp. 4369--4376.

\bibitem{DBLP:journals/tci/XuJWLSZZ20}
S.~Xu, L.~Ji, Z.~Wang, P.~Li, K.~Sun, C.~Zhang, and J.~Zhang, ``Towards reducing severe defocus spread effects for multi-focus image fusion via an optimization based strategy,'' \emph{{IEEE} Transactions Computational Imaging}, vol.~6, pp. 1561--1570, 2020.

\bibitem{DBLP:journals/tgrs/XuALZZL20}
S.~Xu, O.~Amira, J.~Liu, C.~Zhang, J.~Zhang, and G.~Li, ``{HAM-MFN:} hyperspectral and multispectral image multiscale fusion network with {RAP} loss,'' \emph{IEEE Transactions on Geoscience and Remote Sensing}, vol.~58, no.~7, pp. 4618--4628, 2020.

\bibitem{DBLP:journals/inffus/LiZHWC20}
Y.~Li, H.~Zhao, Z.~Hu, Q.~Wang, and Y.~Chen, ``Ivfusenet: Fusion of infrared and visible light images for depth prediction,'' \emph{Information Fusion}, vol.~58, pp. 1--12, 2020.

\bibitem{DBLP:journals/tcsv/MaikCSP07}
V.~Maik, D.~Cho, J.~Shin, and J.~K. Paik, ``Regularized restoration using image fusion for digital auto-focusing,'' \emph{IEEE Transactions on Circuits and Systems for Video Technology}, vol.~17, no.~10, pp. 1360--1369, 2007.

\bibitem{DBLP:journals/inffus/MaCLH16}
J.~Ma, C.~Chen, C.~Li, and J.~Huang, ``Infrared and visible image fusion via gradient transfer and total variation minimization,'' \emph{Information Fusion}, vol.~31, pp. 100--109, 2016.

\bibitem{DBLP:conf/icip/LahoudS18}
F.~Lahoud and S.~S{\"{u}}sstrunk, ``Ar in {VR:} simulating infrared augmented vision,'' in \emph{IEEE International Conference on Image Processing (ICIP)}.\hskip 1em plus 0.5em minus 0.4em\relax {IEEE}, 2018, pp. 3893--3897.

\bibitem{DBLP:conf/iros/HaWKUH17}
Q.~Ha, K.~Watanabe, T.~Karasawa, Y.~Ushiku, and T.~Harada, ``Mfnet: Towards real-time semantic segmentation for autonomous vehicles with multi-spectral scenes,'' in \emph{{IROS}}.\hskip 1em plus 0.5em minus 0.4em\relax {IEEE}, 2017, pp. 5108--5115.

\bibitem{DBLP:journals/inffus/TangYM22}
L.~Tang, J.~Yuan, and J.~Ma, ``Image fusion in the loop of high-level vision tasks: {A} semantic-aware real-time infrared and visible image fusion network,'' \emph{Information Fusion}, vol.~82, pp. 28--42, 2022.

\bibitem{DBLP:journals/corr/abs-2308-02097}
J.~Liu, Z.~Liu, G.~Wu, L.~Ma, R.~Liu, W.~Zhong, Z.~Luo, and X.~Fan, ``Multi-interactive feature learning and a full-time multi-modality benchmark for image fusion and segmentation,'' \emph{CoRR}, vol. abs/2308.02097, 2023.

\bibitem{li2024object}
Z.~Li, Z.~Yuan, L.~Li, D.~Liu, X.~Tang, and F.~Wu, ``Object segmentation-assisted inter prediction for versatile video coding,'' \emph{arXiv preprint arXiv:2403.11694}, 2024.

\bibitem{DBLP:journals/inffus/CaoGHYCQ19}
Y.~Cao, D.~Guan, W.~Huang, J.~Yang, Y.~Cao, and Y.~Qiao, ``Pedestrian detection with unsupervised multispectral feature learning using deep neural networks,'' \emph{Information Fusion}, vol.~46, pp. 206--217, 2019.

\bibitem{DBLP:conf/cvpr/LiuFHWLZL22}
J.~Liu, X.~Fan, Z.~Huang, G.~Wu, R.~Liu, W.~Zhong, and Z.~Luo, ``Target-aware dual adversarial learning and a multi-scenario multi-modality benchmark to fuse infrared and visible for object detection,'' in \emph{Proceedings of the IEEE/CVF Conference on Computer Vision and Pattern Recognition (CVPR)}.\hskip 1em plus 0.5em minus 0.4em\relax {Computer Vision Foundation / IEEE}, 2022, pp. 5792--5801.

\bibitem{DBLP:conf/eccv/LiZHTW18}
C.~Li, C.~Zhu, Y.~Huang, J.~Tang, and L.~Wang, ``Cross-modal ranking with soft consistency and noisy labels for robust {RGB-T} tracking,'' in \emph{{ECCV} {(13)}}, ser. Lecture Notes in Computer Science, vol. 11217.\hskip 1em plus 0.5em minus 0.4em\relax Springer, 2018, pp. 831--847.

\bibitem{DBLP:conf/cvpr/LuWLZLCY20}
Y.~Lu, Y.~Wu, B.~Liu, T.~Zhang, B.~Li, Q.~Chu, and N.~Yu, ``Cross-modality person re-identification with shared-specific feature transfer,'' in \emph{Proceedings of the IEEE/CVF Conference on Computer Vision and Pattern Recognition (CVPR)}.\hskip 1em plus 0.5em minus 0.4em\relax {Computer Vision Foundation / IEEE}, 2020, pp. 13\,376--13\,386.

\bibitem{DBLP:conf/iconip/HarisSU21}
M.~Haris, G.~Shakhnarovich, and N.~Ukita, ``Task-driven super resolution: Object detection in low-resolution images,'' in \emph{{ICONIP} {(5)}}, ser. Communications in Computer and Information Science, vol. 1516.\hskip 1em plus 0.5em minus 0.4em\relax Springer, 2021, pp. 387--395.

\bibitem{DBLP:conf/eccv/PeiHZLW18}
Y.~Pei, Y.~Huang, Q.~Zou, Y.~Lu, and S.~Wang, ``Does haze removal help cnn-based image classification?'' in \emph{{ECCV} {(10)}}, ser. Lecture Notes in Computer Science, vol. 11214.\hskip 1em plus 0.5em minus 0.4em\relax Springer, 2018, pp. 697--712.

\bibitem{DBLP:conf/cvpr/LiARWTJCZGC19}
S.~Li, I.~B. Araujo, W.~Ren, Z.~Wang, E.~K. Tokuda, R.~H. Junior, R.~M.~C. Junior, J.~Zhang, X.~Guo, and X.~Cao, ``Single image deraining: {A} comprehensive benchmark analysis,'' in \emph{Proceedings of the IEEE/CVF Conference on Computer Vision and Pattern Recognition (CVPR)}.\hskip 1em plus 0.5em minus 0.4em\relax {Computer Vision Foundation / IEEE}, 2019, pp. 3838--3847.

\bibitem{DBLP:journals/tmm/ZhouWZML23}
H.~Zhou, W.~Wu, Y.~Zhang, J.~Ma, and H.~Ling, ``Semantic-supervised infrared and visible image fusion via a dual-discriminator generative adversarial network,'' \emph{IEEE Transactions on Multimedia}, vol.~25, pp. 635--648, 2023.

\bibitem{DBLP:journals/inffus/LiWK21}
H.~Li, X.~Wu, and J.~Kittler, ``Rfn-nest: An end-to-end residual fusion network for infrared and visible images,'' \emph{Information Fusion}, vol.~73, pp. 72--86, 2021.

\bibitem{DBLP:journals/tip/LiW19}
H.~Li and X.~Wu, ``Densefuse: {A} fusion approach to infrared and visible images,'' \emph{IEEE Transactions on Image Processing}, vol.~28, no.~5, pp. 2614--2623, 2019.

\bibitem{DBLP:journals/inffus/ZhangLSYZZ20}
Y.~Zhang, Y.~Liu, P.~Sun, H.~Yan, X.~Zhao, and L.~Zhang, ``{IFCNN:} {A} general image fusion framework based on convolutional neural network,'' \emph{Information Fusion}, vol.~54, pp. 99--118, 2020.

\bibitem{zhao2023metafusion}
W.~Zhao, S.~Xie, F.~Zhao, Y.~He, and H.~Lu, ``Metafusion: Infrared and visible image fusion via meta-feature embedding from object detection,'' in \emph{Proceedings of the IEEE/CVF Conference on Computer Vision and Pattern Recognition}, 2023, pp. 13\,955--13\,965.

\bibitem{DBLP:conf/aaai/VinogradovaDM20}
K.~Vinogradova, A.~Dibrov, and G.~Myers, ``Towards interpretable semantic segmentation via gradient-weighted class activation mapping (student abstract),'' in \emph{Proceedings of the AAAI conference on artificial intelligence (AAAI)}, 2020, pp. 13\,943--13\,944.

\bibitem{DBLP:journals/corr/abs-2211-12108}
A.~Kirchknopf, D.~Slijepcevic, I.~Wunderlich, M.~Breiter, J.~Traxler, and M.~Zeppelzauer, ``Explaining {YOLO:} leveraging grad-cam to explain object detections,'' \emph{CoRR}, vol. abs/2211.12108, 2022.

\bibitem{DBLP:conf/cvpr/ShenGTZ20}
Y.~Shen, J.~Gu, X.~Tang, and B.~Zhou, ``Interpreting the latent space of gans for semantic face editing,'' in \emph{Proceedings of the IEEE/CVF Conference on Computer Vision and Pattern Recognition (CVPR)}.\hskip 1em plus 0.5em minus 0.4em\relax {Computer Vision Foundation / IEEE}, 2020, pp. 9240--9249.

\bibitem{DBLP:conf/iclr/BauZSZTFT19}
D.~Bau, J.~Zhu, H.~Strobelt, B.~Zhou, J.~B. Tenenbaum, W.~T. Freeman, and A.~Torralba, ``{GAN} dissection: Visualizing and understanding generative adversarial networks,'' in \emph{Proceedings of theInternational Conference on Learning Representations (ICLR)}.\hskip 1em plus 0.5em minus 0.4em\relax OpenReview.net, 2019.

\bibitem{DBLP:journals/corr/SimonyanVZ13}
K.~Simonyan, A.~Vedaldi, and A.~Zisserman, ``Deep inside convolutional networks: Visualising image classification models and saliency maps,'' in \emph{Proceedings of the International Conference on Learning Representations (ICLR) Workshop}, 2014.

\bibitem{DBLP:journals/corr/SpringenbergDBR14}
J.~T. Springenberg, A.~Dosovitskiy, T.~Brox, and M.~A. Riedmiller, ``Striving for simplicity: The all convolutional net,'' in \emph{Proceedings of the International Conference on Learning Representations (ICLR) Workshop}, 2015.

\bibitem{DBLP:conf/eccv/ZeilerF14}
M.~D. Zeiler and R.~Fergus, ``Visualizing and understanding convolutional networks,'' in \emph{Proceedings of the European Conference on Computer Vision (ECCV)}, ser. Lecture Notes in Computer Science, vol. 8689.\hskip 1em plus 0.5em minus 0.4em\relax Springer, 2014, pp. 818--833.

\bibitem{DBLP:conf/icml/SundararajanTY17}
M.~Sundararajan, A.~Taly, and Q.~Yan, ``Axiomatic attribution for deep networks,'' in \emph{{ICML}}, ser. Proceedings of the European Conference on Computer Vision (ECCV), vol.~70.\hskip 1em plus 0.5em minus 0.4em\relax {PMLR}, 2017, pp. 3319--3328.

\bibitem{DBLP:conf/icml/GregorL10}
K.~Gregor and Y.~LeCun, ``Learning fast approximations of sparse coding,'' in \emph{Proceedings of the International Conference on Machine Learning (ICML)}, J.~F{\"{u}}rnkranz and T.~Joachims, Eds.\hskip 1em plus 0.5em minus 0.4em\relax Omnipress, 2010, pp. 399--406.

\bibitem{DBLP:journals/pami/YangSLX20}
Y.~Yang, J.~Sun, H.~Li, and Z.~Xu, ``Admm-csnet: {A} deep learning approach for image compressive sensing,'' \emph{IEEE Transactions on Pattern Analysis and Machine Intelligence}, vol.~42, no.~3, pp. 521--538, 2020.

\bibitem{DBLP:journals/tci/LiTGME20}
Y.~Li, M.~Tofighi, J.~Geng, V.~Monga, and Y.~C. Eldar, ``Efficient and interpretable deep blind image deblurring via algorithm unrolling,'' \emph{IEEE Transactions on Image Processing}, vol.~6, pp. 666--681, 2020.

\bibitem{DBLP:journals/spm/MongaLE21}
V.~Monga, Y.~Li, and Y.~C. Eldar, ``Algorithm unrolling: Interpretable, efficient deep learning for signal and image processing,'' \emph{IEEE Signal Processing Magazine}, vol.~38, no.~2, pp. 18--44, 2021.

\bibitem{wang2020model}
N.~Wang and J.~Sun, ``Model meets deep learning in image inverse problems,'' \emph{Learning}, vol.~2, no.~9, p.~10, 2020.

\bibitem{DBLP:journals/ijcv/ZhouYPRXC23}
M.~Zhou, K.~Yan, J.~Pan, W.~Ren, Q.~Xie, and X.~Cao, ``Memory-augmented deep unfolding network for guided image super-resolution,'' \emph{International Journal of Computer Vision}, vol. 131, no.~1, pp. 215--242, 2023.

\bibitem{DBLP:conf/mm/SongCZ21}
J.~Song, B.~Chen, and J.~Zhang, ``Memory-augmented deep unfolding network for compressive sensing,'' in \emph{Proceedings of the ACM International Conference on Multimedia (ACM MM)}.\hskip 1em plus 0.5em minus 0.4em\relax {ACM}, 2021, pp. 4249--4258.

\bibitem{DBLP:journals/sigpro/LiuJWSD14}
Y.~Liu, J.~Jin, Q.~Wang, Y.~Shen, and X.~Dong, ``Region level based multi-focus image fusion using quaternion wavelet and normalized cut,'' \emph{Signal Processing}, vol.~97, pp. 9--30, 2014.

\bibitem{DBLP:journals/ijon/LiuMD17}
X.~Liu, W.~Mei, and H.~Du, ``Structure tensor and nonsubsampled shearlet transform based algorithm for {CT} and {MRI} image fusion,'' \emph{Neurocomputing}, vol. 235, pp. 131--139, 2017.

\bibitem{DBLP:journals/isci/0019LLM020}
J.~Chen, X.~Li, L.~Luo, X.~Mei, and J.~Ma, ``Infrared and visible image fusion based on target-enhanced multiscale transform decomposition,'' \emph{Information Sciences}, vol. 508, pp. 64--78, 2020.

\bibitem{DBLP:journals/tip/LiWK20}
H.~Li, X.~Wu, and J.~Kittler, ``Mdlatlrr: {A} novel decomposition method for infrared and visible image fusion,'' \emph{IEEE Transactions on Image Processing}, vol.~29, pp. 4733--4746, 2020.

\bibitem{DBLP:journals/spl/LiuCWW16}
Y.~Liu, X.~Chen, R.~K. Ward, and Z.~J. Wang, ``Image fusion with convolutional sparse representation,'' \emph{IEEE Signal Processing Letters}, vol.~23, no.~12, pp. 1882--1886, 2016.

\bibitem{DBLP:conf/icip/CvejicLBC06}
N.~Cvejic, J.~J. Lewis, D.~R. Bull, and C.~N. Canagarajah, ``Region-based multimodal image fusion using {ICA} bases,'' in \emph{IEEE International Conference on Image Processing (ICIP)}.\hskip 1em plus 0.5em minus 0.4em\relax {IEEE}, 2006, pp. 1801--1804.

\bibitem{BULANON200912}
D.~Bulanon, T.~Burks, and V.~Alchanatis, ``Image fusion of visible and thermal images for fruit detection,'' \emph{Biosystems Engineering}, vol. 103, no.~1, pp. 12 -- 22, 2009.

\bibitem{JIN20181}
X.~Jin, Q.~Jiang, S.~Yao, D.~Zhou, R.~Nie, S.-J. Lee, and K.~He, ``Infrared and visual image fusion method based on discrete cosine transform and local spatial frequency in discrete stationary wavelet transform domain,'' \emph{Infrared Physics \& Technology}, vol.~88, pp. 1 -- 12, 2018.

\bibitem{DBLP:journals/mta/EKR19}
V.~E, M.~K, and S.~B. R, ``Multifocus image fusion scheme based on discrete cosine transform and spatial frequency,'' \emph{Multimedia Tools and Applications}, vol.~78, no.~13, pp. 17\,573--17\,587, 2019.

\bibitem{XIANG201553}
T.~Xiang, L.~Yan, and R.~Gao, ``A fusion algorithm for infrared and visible images based on adaptive dual-channel unit-linking pcnn in nsct domain,'' \emph{Infrared Physics \& Technology}, vol.~69, pp. 53 -- 61, 2015.

\bibitem{DBLP:journals/inffus/HuL12}
J.~Hu and S.~Li, ``The multiscale directional bilateral filter and its application to multisensor image fusion,'' \emph{Information Fusion}, vol.~13, no.~3, pp. 196--206, 2012.

\bibitem{DBLP:conf/icip/ZhangWMW04}
J.~Zhang, L.~Wei, Q.~Miao, and Y.~J. Wang, ``Image fusion based on non-negative matrix factorization,'' in \emph{IEEE International Conference on Image Processing (ICIP)}.\hskip 1em plus 0.5em minus 0.4em\relax {IEEE}, 2004, pp. 973--976.

\bibitem{DBLP:conf/aaai/Xu0LJG20}
H.~Xu, J.~Ma, Z.~Le, J.~Jiang, and X.~Guo, ``Fusiondn: {A} unified densely connected network for image fusion,'' in \emph{Proceedings of the AAAI conference on artificial intelligence (AAAI)}.\hskip 1em plus 0.5em minus 0.4em\relax Proceedings of the AAAI conference on artificial intelligence (AAAI) Press, 2020, pp. 12\,484--12\,491.

\bibitem{DBLP:journals/tim/LiWD20}
H.~Li, X.~Wu, and T.~S. Durrani, ``Nestfuse: An infrared and visible image fusion architecture based on nest connection and spatial/channel attention models,'' \emph{IEEE Transactions on Instrumentation and Measurement}, vol.~69, no.~12, pp. 9645--9656, 2020.

\bibitem{DBLP:conf/ijcai/ZhaoXZLZL20}
Z.~Zhao, S.~Xu, C.~Zhang, J.~Liu, J.~Zhang, and P.~Li, ``Didfuse: Deep image decomposition for infrared and visible image fusion,'' in \emph{Proceedings of the Thirtieth International Joint Conference on Artificial Intelligence (IJCAI)}.\hskip 1em plus 0.5em minus 0.4em\relax ijcai.org, 2020, pp. 970--976.

\bibitem{DBLP:journals/inffus/MaYLLJ19}
J.~Ma, W.~Yu, P.~Liang, C.~Li, and J.~Jiang, ``Fusiongan: {A} generative adversarial network for infrared and visible image fusion,'' \emph{Information Fusion}, vol.~48, pp. 11--26, 2019.

\bibitem{DBLP:journals/inffus/MaLYCGWJ20}
J.~Ma, P.~Liang, W.~Yu, C.~Chen, X.~Guo, J.~Wu, and J.~Jiang, ``Infrared and visible image fusion via detail preserving adversarial learning,'' \emph{Information Fusion}, vol.~54, pp. 85--98, 2020.

\bibitem{DBLP:journals/tip/MaXJMZ20}
J.~Ma, H.~Xu, J.~Jiang, X.~Mei, and X.~S. Zhang, ``Ddcgan: {A} dual-discriminator conditional generative adversarial network for multi-resolution image fusion,'' \emph{IEEE Transactions on Image Processing}, vol.~29, pp. 4980--4995, 2020.

\bibitem{DBLP:journals/tcsv/ZhaoXZLZL22}
Z.~Zhao, S.~Xu, J.~Zhang, C.~Liang, C.~Zhang, and J.~Liu, ``Efficient and model-based infrared and visible image fusion via algorithm unrolling,'' \emph{IEEE Transactions on Circuits and Systems for Video Technology}, vol.~32, no.~3, pp. 1186--1196, 2022.

\bibitem{DBLP:journals/pami/0002D21}
X.~Deng and P.~L. Dragotti, ``Deep convolutional neural network for multi-modal image restoration and fusion,'' \emph{IEEE Transactions on Pattern Analysis and Machine Intelligence}, vol.~43, no.~10, pp. 3333--3348, 2021.

\bibitem{DBLP:journals/corr/abs-2005-08448}
Z.~Zhao, J.~Zhang, H.~Bai, Y.~Wang, Y.~Cui, L.~Deng, K.~Sun, C.~Zhang, J.~Liu, and S.~Xu, ``Deep convolutional sparse coding networks for interpretable image fusion,'' in \emph{Proceedings of the IEEE/CVF Conference on Computer Vision and Pattern Recognition (CVPR) Workshops}.\hskip 1em plus 0.5em minus 0.4em\relax {Computer Vision Foundation / IEEE}, 2023, pp. 2369--2377.

\bibitem{DBLP:journals/pami/XuMJGL22}
H.~Xu, J.~Ma, J.~Jiang, X.~Guo, and H.~Ling, ``U2fusion: {A} unified unsupervised image fusion network,'' \emph{IEEE Transactions on Pattern Analysis and Machine Intelligence}, vol.~44, no.~1, pp. 502--518, 2022.

\bibitem{DBLP:journals/ijcv/ZhangM21}
H.~Zhang and J.~Ma, ``Sdnet: {A} versatile squeeze-and-decomposition network for real-time image fusion,'' \emph{International Journal of Computer Vision}, vol. 129, no.~10, pp. 2761--2785, 2021.

\bibitem{DBLP:conf/eccv/LiangJLM22}
P.~Liang, J.~Jiang, X.~Liu, and J.~Ma, ``Fusion from decomposition: {A} self-supervised decomposition approach for image fusion,'' in \emph{Proceedings of the European Conference on Computer Vision (ECCV)}, ser. Lecture Notes in Computer Science, vol. 13678.\hskip 1em plus 0.5em minus 0.4em\relax Springer, 2022, pp. 719--735.

\bibitem{DBLP:conf/eccv/HuangLFLZL22}
Z.~Huang, J.~Liu, X.~Fan, R.~Liu, W.~Zhong, and Z.~Luo, ``Reconet: Recurrent correction network for fast and efficient multi-modality image fusion,'' in \emph{Proceedings of the European Conference on Computer Vision (ECCV)}, ser. Lecture Notes in Computer Science, vol. 13678.\hskip 1em plus 0.5em minus 0.4em\relax Springer, 2022, pp. 539--555.

\bibitem{DBLP:conf/ijcai/WangLFL22}
D.~Wang, J.~Liu, X.~Fan, and R.~Liu, ``Unsupervised misaligned infrared and visible image fusion via cross-modality image generation and registration,'' in \emph{Proceedings of the Thirtieth International Joint Conference on Artificial Intelligence (IJCAI)}.\hskip 1em plus 0.5em minus 0.4em\relax ijcai.org, 2022, pp. 3508--3515.

\bibitem{DBLP:conf/cvpr/Xu0YLL22}
H.~Xu, J.~Ma, J.~Yuan, Z.~Le, and W.~Liu, ``Rfnet: Unsupervised network for mutually reinforcing multi-modal image registration and fusion,'' in \emph{Proceedings of the IEEE/CVF Conference on Computer Vision and Pattern Recognition (CVPR)}.\hskip 1em plus 0.5em minus 0.4em\relax {Computer Vision Foundation / IEEE}, 2022, pp. 19\,647--19\,656.

\bibitem{DBLP:conf/cvpr/ZhaoZXLP22}
Z.~Zhao, J.~Zhang, S.~Xu, Z.~Lin, and H.~Pfister, ``Discrete cosine transform network for guided depth map super-resolution,'' in \emph{Proceedings of the IEEE/CVF Conference on Computer Vision and Pattern Recognition (CVPR)}, 2022, pp. 5687--5697.

\bibitem{deng2019deep}
X.~Deng and P.~L. Dragotti, ``Deep coupled ista network for multi-modal image super-resolution,'' \emph{IEEE Transactions on Image Processing}, vol.~29, pp. 1683--1698, 2019.

\bibitem{DBLP:journals/tip/MarivaniTCD20}
I.~Marivani, E.~Tsiligianni, B.~Cornelis, and N.~Deligiannis, ``Multimodal deep unfolding for guided image super-resolution,'' \emph{IEEE Transactions on Image Processing}, vol.~29, pp. 8443--8456, 2020.

\bibitem{yang2022memory}
G.~Yang, M.~Zhou, K.~Yan, A.~Liu, X.~Fu, and F.~Wang, ``Memory-augmented deep conditional unfolding network for pan-sharpening,'' in \emph{Proceedings of the IEEE/CVF Conference on Computer Vision and Pattern Recognition}, 2022, pp. 1788--1797.

\bibitem{DBLP:conf/cvpr/ZhangGT20}
K.~Zhang, L.~V. Gool, and R.~Timofte, ``Deep unfolding network for image super-resolution,'' in \emph{Proceedings of the IEEE/CVF Conference on Computer Vision and Pattern Recognition (CVPR)}, 2020, pp. 3214--3223.

\bibitem{DBLP:conf/cvpr/ZhangZGZ17}
K.~Zhang, W.~Zuo, S.~Gu, and L.~Zhang, ``Learning deep {CNN} denoiser prior for image restoration,'' in \emph{Proceedings of the IEEE/CVF Conference on Computer Vision and Pattern Recognition (CVPR)}, 2017, pp. 2808--2817.

\bibitem{DBLP:journals/pami/DongWYSWL19}
W.~Dong, P.~Wang, W.~Yin, G.~Shi, F.~Wu, and X.~Lu, ``Denoising prior driven deep neural network for image restoration,'' \emph{IEEE Transactions on Pattern Analysis and Machine Intelligence}, vol.~41, no.~10, pp. 2305--2318, 2019.

\bibitem{DBLP:conf/iccv/LiuPRS19}
Y.~Liu, J.~Pan, J.~S.~J. Ren, and Z.~Su, ``Learning deep priors for image dehazing,'' in \emph{{ICCV}}.\hskip 1em plus 0.5em minus 0.4em\relax {IEEE} Computer Society, 2019, pp. 2492--2500.

\bibitem{DBLP:conf/cvpr/Xu0ZSL021}
S.~Xu, J.~Zhang, Z.~Zhao, K.~Sun, J.~Liu, and C.~Zhang, ``Deep gradient projection networks for pan-sharpening,'' in \emph{Proceedings of the IEEE/CVF Conference on Computer Vision and Pattern Recognition (CVPR)}, 2021, pp. 1366--1375.

\bibitem{li2023lrrnet}
H.~Li, T.~Xu, X.-J. Wu, J.~Lu, and J.~Kittler, ``Lrrnet: A novel representation learning guided fusion network for infrared and visible images,'' \emph{IEEE transactions on pattern analysis and machine intelligence}, 2023.

\bibitem{DBLP:conf/cvpr/ZhouKLOT16}
B.~Zhou, A.~Khosla, {\`{A}}.~Lapedriza, A.~Oliva, and A.~Torralba, ``Learning deep features for discriminative localization,'' in \emph{Proceedings of the IEEE/CVF Conference on Computer Vision and Pattern Recognition (CVPR)}.\hskip 1em plus 0.5em minus 0.4em\relax {Computer Vision Foundation / IEEE}, 2016, pp. 2921--2929.

\bibitem{DBLP:conf/iccv/SelvarajuCDVPB17}
R.~R. Selvaraju, M.~Cogswell, A.~Das, R.~Vedantam, D.~Parikh, and D.~Batra, ``Grad-cam: Visual explanations from deep networks via gradient-based localization,'' in \emph{Proceedings of the IEEE International Conference on Computer Vision (ICCV)}.\hskip 1em plus 0.5em minus 0.4em\relax {IEEE} Computer Society, 2017, pp. 618--626.

\bibitem{DBLP:journals/corr/abs-1805-11393}
S.~Lee, J.~Lee, J.~Lee, C.~Park, and S.~Yoon, ``Robust tumor localization with pyramid grad-cam,'' \emph{CoRR}, vol. abs/1805.11393, 2018.

\bibitem{DBLP:journals/jmlr/BaehrensSHKHM10}
D.~Baehrens, T.~Schroeter, S.~Harmeling, M.~Kawanabe, K.~Hansen, and K.~M{\"{u}}ller, ``How to explain individual classification decisions,'' \emph{Journal of Machine Learning Research}, vol.~11, pp. 1803--1831, 2010.

\bibitem{DBLP:conf/cvpr/GuD21}
J.~Gu and C.~Dong, ``Interpreting super-resolution networks with local attribution maps,'' in \emph{Proceedings of the IEEE/CVF Conference on Computer Vision and Pattern Recognition (CVPR)}.\hskip 1em plus 0.5em minus 0.4em\relax {Computer Vision Foundation / IEEE}, 2021, pp. 9199--9208.

\bibitem{DBLP:journals/ijgt/Friedman04}
E.~J. Friedman, ``Paths and consistency in additive cost sharing,'' \emph{International Journal of Game Theory}, vol.~32, no.~4, pp. 501--518, 2004.

\bibitem{DBLP:conf/eccv/ChenZPSA18}
L.~Chen, Y.~Zhu, G.~Papandreou, F.~Schroff, and H.~Adam, ``Encoder-decoder with atrous separable convolution for semantic image segmentation,'' in \emph{Proceedings of the European Conference on Computer Vision (ECCV)}, ser. Lecture Notes in Computer Science, vol. 11211.\hskip 1em plus 0.5em minus 0.4em\relax Springer, 2018, pp. 833--851.

\bibitem{DBLP:journals/inffus/TangYZJM22}
L.~Tang, J.~Yuan, H.~Zhang, X.~Jiang, and J.~Ma, ``Piafusion: {A} progressive infrared and visible image fusion network based on illumination aware,'' \emph{Infromation Fusion}, vol. 83-84, pp. 79--92, 2022.

\bibitem{zhao2023ddfm}
Z.~Zhao, H.~Bai, Y.~Zhu, J.~Zhang, S.~Xu, Y.~Zhang, K.~Zhang, D.~Meng, R.~Timofte, and L.~Van~Gool, ``Ddfm: Denoising diffusion model for multi-modality image fusion,'' \emph{arXiv e-prints}, pp. arXiv--2303, 2023.

\bibitem{xu2023murf}
H.~Xu, J.~Yuan, and J.~Ma, ``Murf: Mutually reinforcing multi-modal image registration and fusion,'' \emph{IEEE Transactions on Pattern Analysis and Machine Intelligence}, 2023.

\bibitem{yi2024diff}
X.~Yi, L.~Tang, H.~Zhang, H.~Xu, and J.~Ma, ``Diff-if: Multi-modality image fusion via diffusion model with fusion knowledge prior,'' \emph{Information Fusion}, vol. 110, p. 102450, 2024.

\bibitem{yi2024text}
X.~Yi, H.~Xu, H.~Zhang, L.~Tang, and J.~Ma, ``Text-if: Leveraging semantic text guidance for degradation-aware and interactive image fusion,'' in \emph{Proceedings of the IEEE/CVF Conference on Computer Vision and Pattern Recognition (CVPR)}, 2024, pp. 27\,026--27\,035.

\bibitem{xie2021segformer}
E.~Xie, W.~Wang, Z.~Yu, A.~Anandkumar, J.~M. Alvarez, and P.~Luo, ``Segformer: Simple and efficient design for semantic segmentation with transformers,'' \emph{Advances in neural information processing systems}, vol.~34, pp. 12\,077--12\,090, 2021.

\end{thebibliography}

\begin{IEEEbiographynophoto}{Haowen Bai}
is pursuing a Ph.D. degree in statistics at the School of Mathematics and Statistics, Xi'an Jiaotong University, Xi'an, China. His research interests include computer vision, machine learning, deep learning, and especially in image/video restoration, low-level vision, and computational imaging.
\end{IEEEbiographynophoto}

\begin{IEEEbiographynophoto}{Zixiang Zhao}
is currently a postdoctoral researcher at the Photogrammetry and Remote Sensing Group, ETH Zürich, Switzerland. He received his Ph.D. degree in statistics from the School of Mathematics and Statistics, Xi’an Jiaotong University, Xi’an, China. Previously, he was a visiting Ph.D. student at the Computer Vision Lab, ETH Zürich, Switzerland, and also worked as a research assistant at the Visual Computing Group, Harvard University, USA. His research interests include computer vision, machine learning, deep learning, with a particular focus on image/video restoration, low-level vision, and computational imaging.
\end{IEEEbiographynophoto}

\begin{IEEEbiographynophoto}{Jiangshe Zhang}
received the B.S., M.S., and Ph.D. degrees in applied mathematics from Xi'an Jiaotong University, Xi'an, China, in 1984, 1987, and 1993, respectively.
He is currently a Professor with the School of Mathematics and Statistics, Xi'an Jiaotong University. He has authored and coauthored one monograph and more than 150 journal papers. His current research interests include fuzzy sets and logic, machine learning, deep learning, and statistical decision making.
Dr. Zhang was the recipient of the Second Prize of National Natural Science Award of China (Third Place) in 2007. He was a Member of the academic committee of Tianyuan Mathematical Center in Northwest China and the President of Statistics Society of the Shaanxi Mathematical Society.
\end{IEEEbiographynophoto}

\begin{IEEEbiographynophoto}{Baisong Jiang}
is pursuing a Ph.D. degree in statistics at the School of Mathematics and Statistics, Xi'an Jiaotong University, Xi'an, China.  His research interests include machine learning, deep learning, and especially in image restoration, and seismic data reconstruction.
\end{IEEEbiographynophoto}

\begin{IEEEbiographynophoto}{Lilun Deng}
is pursuing a Ph.D. degree in statistics at the School of Mathematics and Statistics, Xi'an Jiaotong University, Xi'an, China. His research interests include multi-modal image fusion, semantic segmentation, and artificial intelligence for exploration geophysics.
\end{IEEEbiographynophoto}

\begin{IEEEbiographynophoto}{Yukun Cui}
is pursuing a Ph.D. degree in mathematics at the School of Mathematics and Statistics, Xi'an Jiaotong University, Xi'an, China. His research interests include deep learning, machine learning, computer vision, and robust optimization.
\end{IEEEbiographynophoto}

\begin{IEEEbiographynophoto}{Shuang Xu}
received the Ph.D. degree in statistics from Xi’an Jiaotong University, Xi’an, China, in 2021.
He is currently working with the School of Mathematics and Statistics, Northwestern Polytechnical University, Xi’an. His research interests include Bayesian statistics, deep learning, and complex network and system.
\end{IEEEbiographynophoto}

\begin{IEEEbiographynophoto}{Chunxia Zhang}
received the Ph.D. degree in applied mathematics from Xi’an Jiaotong University, Xi’an, China, in 2010. She is currently a Professor with the School of Mathematics and Statistics, Xi’an Jiaotong University. She has authored and coauthored about 30 journal articles on ensemble learning techniques, nonparametric regression, and so on. Her main interests are in the area of ensemble learning, variable selection, and deep learning.
\end{IEEEbiographynophoto}

\vfill
\end{document}